%% file: main.tex
\documentclass[sigconf]{acmart}
\usepackage{times}
\usepackage{amsmath,amssymb,amsthm,amsfonts}
\usepackage{algorithm,algpseudocode}
\hypersetup{breaklinks=true}
\usepackage{multirow}
\usepackage{caption}
\usepackage{subcaption}

\usepackage{tikz}
\usetikzlibrary{arrows.meta}
\tikzset{%
  >={Latex[width=2mm,length=2mm]},
            base/.style = {rectangle, rounded corners, draw=black,
                          minimum width=0.5cm, minimum height=1cm,
                          text centered, font=\sffamily},
  sanitizerNode/.style = {base, fill=blue!30},
      discriminatorNode/.style = {base, fill=red!30},
    originals/.style = {base, fill=green!30, font=\ttfamily},
         process/.style = {base, minimum width=0.5cm, fill=orange!15, font=\ttfamily},
          costs/.style = {base, minimum width=0.5cm, fill=black!15, font=\ttfamily},
          noiseNodes/.style = {base, minimum width=0.5cm, fill=black!50, font=\ttfamily},
}
\input{Macros}
\title{Adversarial training approach for local data debiasing}

\renewcommand\footnotetextcopyrightpermission[1]{} 

\setcopyright{none}
\begin{document}
    \input{Authors}
    \input{01-Abstract}
    \maketitle
    \pagestyle{plain} 
    \input{02-Introduction}
    \input{03-SystemModel}
    \input{04-FairnessDef}
    \input{05-RelatedWork}
    \input{07-TheProposedFramework}

    \input{08-Experiments}

    \input{09-Conclusion-FutureWork}
    \input{myBib}
    \section*{Appendices}
    \input{10-Appendices}
    \input{11-Appendices-german}

\end{document}

%% file: Macros.tex
\newcommand{\dataset}{$D$}
\newcommand{\records}{\textit{r}}
\newcommand{\sensAttr}{\textit{S}}
\newcommand{\sensAttrL}{\textit{s}}

\newcommand{\decAttr}{Y}
\newcommand{\decAttrL}{y}

\newcommand{\notSensAttr}{\textit{A}}

\newcommand{\dis}{$D_{isc}$}
\newcommand{\san}{$S_{an}$}

\def\rm#1{\textcolor{gray}{}}

\newcommand{\gansan}{\textsf{GANSan\,}}

\newcommand{\myBer}{BER}
\newcommand{\mySac}{sAcc}
\newcommand{\myDI}{DemoParity}
\newcommand{\eqod}{EqOddGap}

\newcommand{\sandist}{fid}
\newcommand{\mdist}{diversity}
\newcommand{\myAcc}{yAcc}
\newcommand{\heuristic}{HeuristicA}

%% file: Authors.tex
\author{Ulrich Aïvodji}
\email{aivodji.ulrich@courrier.uqam.ca}
\affiliation{%
  \institution{Université du Québec à Montréal}
  \streetaddress{405 Sainte-Catherine Street East}
  \city{Montréal}
  \state{Québec}
  \country{Canada}
  \postcode{H2L 2C4}
}

\author{François Bidet}
\email{francois.bidet@polytechnique.edu}
\affiliation{%
  \institution{Ecole Polytechnique}
  \streetaddress{Route de Saclay}
  \city{Palaiseau Cedex}
  \country{France}
  \postcode{91128}
}

\author{Sébastien Gambs}
\email{gambs.sebastien@uqam.ca}
\affiliation{%
  \institution{Université du Québec à Montréal}
  \streetaddress{405 Sainte-Catherine Street East}
  \city{Montréal}
  \state{Québec}
  \country{Canada}
  \postcode{H2L 2C4}
}

\author{Rosin Claude Ngueveu}
\email{ngueveu.rosin_claude@uqam.ca}
\affiliation{%
  \institution{Université du Québec à Montréal}
  \streetaddress{405 Sainte-Catherine Street East}
  \city{Montréal}
  \state{Québec}
  \country{Canada}
  \postcode{H2L 2C4}
}

\author{Alain Tapp}
\email{alain.tapp@gmail.com}
\affiliation{%
  \institution{Université de Montréal}
  \streetaddress{succursale Centre-ville}
  \city{Montréal}
  \state{Québec}
  \country{Canada}
  \postcode{H3C 3J7}
}

\renewcommand{\shortauthors}{Aivodji, Bidet, Gambs, Ngueveu and Tapp}

%% file: 01-Abstract.tex
\begin{abstract}
The widespread use of automated decision processes in many areas of our society raises serious ethical issues with respect to the fairness of the process and the possible resulting discriminations. 
To solve this issue, we propose a novel adversarial training approach called \gansan for learning a sanitizer whose objective is to prevent the possibility of \emph{any} discrimination (\emph{i.e.}, direct and indirect) based on a sensitive attribute by removing the attribute itself as well as the existing correlations with the remaining attributes. 
Our method \gansan is partially inspired by the powerful framework of generative adversarial networks (in particular Cycle-GANs), which offers a flexible way to learn a distribution empirically or to translate between two different distributions.
In contrast to prior work, one of the strengths of our approach is that the sanitization is performed in the same space as the original data by only modifying the other attributes as little as possible, thus preserving the interpretability of the sanitized data. 
Consequently, once the sanitizer is trained, it can be applied to new data locally by an individual on his profile before releasing it. 
Finally, experiments on a real datasets demonstrate the effectiveness of the approach as well as the achievable trade-off between fairness and utility.
\end{abstract} 

%% file: 02-Introduction.tex
\section{Introduction}

In recent years, the availability and the diversity of large-scale datasets, the algorithmic advancements in machine learning and the increase in computational power have led to the development of personalized services and prediction systems to such an extent that their use is now ubiquitous in our society. 
For instance, machine learning-based systems are now used in banking for assessing the risk associated with loan applications~\cite{CredScoPatent}, in hiring system~\cite{FalRecr2012} and in predictive justice to quantify the recidivism risk of an inmate~\cite{Center2016}. 
Despite their usefulness, the predictions performed by these algorithms are not exempt from biases, and numerous cases of discriminatory decisions have been reported over the last years.

For example, going back on the case of predictive justice, a study conducted by ProPublica showed that the recidivism prediction tool COMPAS, which is currently used in Broward County (Florida), is strongly biased against black defendants, by displaying a false positive rate twice as high for black persons than for white persons~\cite{JuliaAng2016}.
If the dataset exhibits strong detectable biases towards a particular sensitive group (\emph{e.g.}, an ethnic or minority group), the naïve solution of removing the attribute identifying the sensitive group prevents only \emph{direct discrimination}.
Indeed, \emph{indirect discrimination} can still occur due to correlations between the sensitive attribute and other attributes.

In this paper, we propose a novel approach called \gansan (for \emph{Generative Adversarial Network Sanitizer}) to address the problem of discrimination due to the biased underlying data.

In a nutshell, our approach learns a sanitizer (in our case a neural network) transforming the input data in a way that maximize the following two metrics : (1) \emph{fidelity}, in the sense that the transformation should modify the data as little as possible, and (2) \emph{non-discrimination}, which means that the sensitive attribute should be difficult to predict from the sanitized data.

A typical use case might be one in which a company during its recruitment process offers to job applicants a tool to remove racial correlation in their data before submitting their sanitized profile on the job application platform. 
If built appropriately, this tool would make the recruitment process of the company free from racial discrimination as it never had access to the original profile.

Overall, our contributions can be summarized as follows.
\begin{itemize}
\item We propose a novel adversarial approach, inspired from Generative Adversarial Networks (GANs) \cite{NIPS2014_5423}, in which a sanitizer is learned from data representing the population. 
The sanitizer can then be applied on a profile in such way that the sensitive attribute is removed, as well as existing correlations with other attributes while ensuring that the sanitized profile is modified as little as possible, preventing both direct and indirect discrimination.

\item Our objective is more generic than simply building a non-discriminating classifier, in the sense that we aim at debiasing the data with respect to the sensitive attribute.
Thus, one of the main benefits of our approach is that the sanitization can be performed without having any knowledge regarding the tasks that are going to be conducted in the future on the sanitized data. 

\item Another strength of our approach is that once the sanitizer has been learned, it can be used locally by an individual (\emph{e.g.}, on a device under his control) to generate a modified version of his profile that still lives in the same representation space, but from which it is very difficult to infer the sensitive attribute.
In this sense, our method can be considered to fall under the category of \emph{randomized response techniques}~\cite{RanResp1965} as it can be distributed before being used locally by a user to sanitize his data. 
Thus, it does not require his true profile to be sent to a trusted third party.
Of all of the approaches that currently exist in the literature to reach algorithmic fairness~\cite{ACompFairEnInter2018}, we are not aware of any other work that has considered the local sanitization with the exception of~\cite{romanelli2019generating}, which focuses on the protection of privacy but could also be applied to enhance fairness.

\item To demonstrate its usefulness, we have proposed and discussed four different evaluation scenarios and assessed our approach on real datasets for these four different scenarios. 
In particular, we have analyzed the achievable trade-off between fairness and utility measured both in terms of the perturbations introduced by the sanitization framework but also with respect to the accuracy of a classifier learned on the sanitized data.
\end{itemize}

The outline of the paper is as follows. 
First, in Section \ref{sec:systemmodel}, we introduce the system model before reviewing the background notions on fairness metrics.
Afterwards, in Section \ref{sec:rel_work}, we review the related work on methods for enhancing fairness belonging to the preprocessing approach like ours before describing \gansan in Section \ref{sec:our-approach}.
Finally, we evaluate experimentally our approach in Section \ref{sec:experiments} before concluding in Section \ref{discuss_conclude}.

%% file: 03-SystemModel.tex
\section{Preliminaries}
\label{sec:systemmodel}

In this section, we first present the system model used in this paper before reviewing the background notions on fairness metrics.

\subsection{System model} 
In this paper, we consider the generic setting of a dataset $D$ composed of $N$ records.
Each record $\records_i$ typically corresponds to the profile of the individual $i$ and is made of $d$ attributes, which can be categorical, discrete or continuous.
Amongst those, the \emph{sensitive attribute} $\sensAttr{}$ (\emph{e.g.}, gender, ethnic origin, religious belief, \ldots) should remain hidden to prevent discrimination. 
In addition, the \emph{decision attribute} $\decAttr$ is typically used for a classification task (\emph{e.g.}, accept or reject an individual for a job interview).
The other attributes of the profile, which are neither $\sensAttr{}$ nor $\decAttr{}$, will be referred hereafter as $\notSensAttr{}$.

For simplicity, in this work we restrict ourselves to the situations in which these two attributes are binary (\emph{i.e.}, $\sensAttr{} \in \{0,1\}$ and $\decAttr \in \{0,1\}$ ).
However, our approach can also be generalized to multivalued attributes, although quantifying fairness for multivalued attributes is much more challenging than for binary ones \cite{kearns2017preventing}.
Our main objective is to prevent the possibility of inferring the sensitive attribute from the sanitized data. 

This objective is similar to the protection against \emph{group membership inference}, which in our context amounts to distinguish between the two groups generated by the values of \sensAttr, which we will refer to as the \emph{sensitive group} (for which $\sensAttr =0$) and the \emph{default group} (for which $\sensAttr =1$).

%% file: 04-FairnessDef.tex
\subsection{Fairness metrics}
\label{sec:fairness}

First, we would like to point out that there are many different definitions of fairness existing in the literature~\cite{dwork2012fairness,joseph2016fairness,corbett2017algorithmic,narayanan21def2018,ACompFairEnInter2018,verma2018fairness} and that the choice of the appropriate definition is highly dependent of the context considered.

For instance, one natural approach for defining fairness is the concept of \emph{individual fairness}~\cite{dwork2012fairness}, which states that individuals that are similar except for the sensitive attribute should be treated similarly (\emph{i.e}, receive similar decisions).
This notion relates to the legal concept of \emph{disparate treatment}~\cite{barocas2016big}, which occurs if the decision process was made based on sensitive attributes.
This definition is relevant when discrimination is caused by the decision process.
Therefore, it cannot be used in the situation in which the objective is to directly redress biases in the data.

In contrast to individual fairness, \emph{group fairness} relies on statistic of outcomes of the subgroups indexed by \sensAttr{} and can be quantified in several ways, such as \emph{demographic parity}~\cite{berk2018fairness} and \emph{equalized odds}~\cite{hardt2016equality}.
More precisely, the demographic parity corresponds to the absolute difference of rates of positive outcomes in the sensitive and default groups (for which respectively $\sensAttr{} = 0$ and $\sensAttr{} = 1$):
\begin{align}
    \myDI{} = |P(\decAttr|\sensAttr{}=0) - P(\decAttr|\sensAttr{}=1)|,
    \label{eq:demopar}
\end{align}
while \emph{equalized odds} is the absolute difference of odds in each subgroup:
\begin{align}
    \begin{split}
        \eqod{}_{\decAttrL{}} = &|Pr(\hat{\decAttr{}} = 1| \sensAttr{}=0, \decAttr{}=\decAttrL{}) \\
                                &- Pr(\hat{\decAttr{}} = 1| \sensAttr{}=1, \decAttr{}=\decAttrL{})|    
        \label{eq:eqod}
    \end{split}
\end{align}

Compared to demographic parity, equalized odds is more suitable when the base rates in both groups differ ($P(\decAttr{}=1|\sensAttr{}=0) \neq P(\decAttr{}=1|\sensAttr{}=1)$).
Note that these definitions are agnostic to the cause of the discrimination and are based solely on the assumption that statistics of outcomes should be similar between subgroups. 

In our work, we follow a different line of research by defining fairness in terms of \emph{the inability to infer \sensAttr{} from other attributes}~\cite{Feldman2014,xu2018fairgan}.
This approach stems from the observation that it is impossible to discriminate based on the sensitive attribute if the latter is unknown and cannot be predicted from other attributes.
Thus, our approach aims at sanitizing the data in such a way that no classifier should be able to infer the sensitive attribute from the sanitized data.

The inability to infer the attribute \sensAttr{} can be measured by the accuracy of a predictor \emph{Adv} trained to recover the hidden \sensAttr{} (\emph{\mySac{}}), as well as the \textit{balanced error rate (\myBer{})} introduced in~\cite{Feldman2014}: 
\begin{align}
\myBer{}(Adv(\notSensAttr, \decAttr), \sensAttrL{}) = \dfrac{1}{2} (\sum_{\sensAttrL{}=0}^{1}P(Adv(\notSensAttr, \decAttr) \neq \sensAttrL{}| \sensAttr{} = \sensAttrL{})).
\end{align}
The BER captures the predictability of both classes and a value of $\dfrac{1}{2}$ can be considered optimal for protecting against inference in the sense that it means that the inferences made by the predictor are not better than a random guess.
In addition, the \myBer{} is more relevant than the accuracy of a classifier $\mySac{}$ at predicting the sensitive attribute for datasets with imbalanced proportions of sensitive and default groups.
Thus, a successful sanitization would lead to a significant drop of the accuracy while raising the BER close to its optimal value of $0.5$.

%% file: 05-RelatedWork.tex
\section{Related work}
\label{sec:rel_work}

In recent years, many approaches have been developed to enhance the fairness of machine learning algorithms.
Most of these techniques can be classified into three families of approaches, namely (1) the \emph{preprocessing approach}~\cite{CRWAA-ALFR2016,Feldman2014,VFAE2015,Zemel2013} in which fairness is achieved by changing the characteristics of the input data (\emph{e.g.} by suppressing undesired correlations with the sensitive attribute), (2) the \emph{algorithmic modification approach} (also sometimes called \emph{constrained optimization}) in which the learning algorithm is adapted to ensure that it is fair by design~\cite{FairnessConsZafar2015,kamishima2012fairness} and (3) the \emph{postprocessing approach} that modifies the output of the learning algorithm to increase the level of fairness~\cite{Kamiran2010,hardt2016equality}\footnote{We refer the interested reader to~\cite{ACompFairEnInter2018} for a recent survey comparing the different fairness-enhancing methods.}.
Due to the limited space and as our approach falls within the preprocessing approach~\cite{CRWAA-ALFR2016,Feldman2014,VFAE2015,Zemel2013} in which fairness is achieved by changing the characteristics of the input data (\emph{e.g.} by suppressing undesired correlations with the sensitive attribute), we will review afterwards only methods of this category that makes use of an adversarial training. 

Several approaches have been explored to enhance fairness based on adversarial learning.
For instance, Edwards and Storkey~\cite{CRWAA-ALFR2016} have trained an encoder to output a representation from which an adversary is unable to predict the group membership, from which a decoder can reconstruct the data and on which decision predictor still performs well.
Madras, Creager, Pitassi and Zemel~\cite{madras2018learning} extended this framework to satisfy the equality of opportunities constraint~\cite{hardt2016equality} and explored the theoretical guarantees for fairness provided by the learned representation as well as the ability of the representation to be used for different classification tasks.
Beutel, Chen, Zhao and Chi~\cite{beutel2017data} have studied the impact of data quality on fairness in the context of adversarial learning, and demonstrated for instance that learning a representation independent of the sensitive attribute with a balanced dataset ensures statistical parity.
Zhang, Lemoine and Mitchell~\cite{zhang2018mitigating} have designed a decision predictor satisfying group fairness by ensuring that an adversary is unable to infer the sensitive attribute from the predicted outcome. 
McNamara, Ong and Williamson~\cite{mcnamara2019costs} have investigated the benefits and drawbacks of fair representation learning, demonstrating that techniques building fair representations restrict the space of possible decisions, limiting the usages of resulting data while providing fairness.

All these previous approaches does not preserve the interpretability of the data, in the sense that the modified profile lives in a different space than the original one.
One notable exception is \textsf{FairGan}~\cite{xu2018fairgan}, which maintains the interpretability of the profile. 
Their objective is to learn a fair classifier on a dataset that has been generated such that it is discrimination-free and whose distribution on attributes is close to the original one.
While \textsf{FairGan} generates a synthetic dataset close to the original data while being discrimination free, one key difference with \gansan is that \textsf{FairGan} cannot be used to sanitize directly a particular profile. 
Following a similar line of work, there is a growing body of research investigating the use of adversarial training to protect the privacy of individuals during the collection or disclosure of data. 
Feutry, Piantanida, Bengio and Duhamel~\cite{feutry2018learning} have proposed an anonymization procedure based on the learning of an encoder, an adversary and a label predictor. 
The authors have ensured the convergence of these three networks during training by proposing an efficient optimization procedure with bounds on the probability of misclassification. 
Pittaluga, Koppal and Chakrabarti~\cite{pittaluga2019learning} have designed a procedure based on adversarial training to hide a private attribute of a dataset. 
Romanelli, Palamidessi and Chatzikokolakis~\cite{romanelli2019generating} have designed a mechanism to create a dataset preserving the original representation. 
They have developed a method for learning an optimal privacy protection mechanism also inspired by GAN~\cite{tripathy2017privacy}, which they have applied to location privacy. 
The objective is to minimize the amount of information (measured by the mutual information) preserved between \sensAttr{} and the prediction made on the decision attribute by a classifier while respecting a bound on utility.
With respect to the local sanitization and \emph{randomized response techniques}, most of them are applied in the context of privacy protection~\cite{wang2016using}. 
Our approach is among the first that places the protection of information at the individual level as the user can locally sanitize his data before publishing it. 

%% file: 07-TheProposedFramework.tex
\section{Adversarial training for data debiasing}
\label{sec:our-approach}

As previously explained, removing the sensitive attribute is rarely sufficient to guarantee non-discrimination as correlations are likely to exist between other attributes and the sensitive one. 

In general, detecting and suppressing complex correlations between attributes is a difficult task.

To address this challenge, our approach \gansan{} relies on the modelling power of GANs 
to build a sanitizer that can cancel out correlations with the sensitive attribute without requiring an explicit model of those correlations.
In particular, it exploits the capacity of the discriminator to distinguish the subgroups indexed by the sensitive attribute.
Once the sanitizer has been trained, any individual can apply it locally on his profile before disclosing it.
The sanitized data can then be safely used for any subsequent task.

\subsection{Generative adversarial network sanitization}

\paragraph{High level overview.}
Formally, given a dataset $D$, the objective of \gansan is to learn a function $S_{an}$, called the \emph{sanitizer} that perturbs individual profiles of the dataset \dataset, such that a distance measure called the fidelity $\sandist$ (in our case we will use the $L_2$ norm) between the original and the sanitized datasets ($\bar{D} = S_{an}(D) = \{\bar{\notSensAttr{}}, \bar{\decAttr{}}\}$), is minimal, while ensuring that \sensAttr{} cannot be recovered from $\bar{D}$. 
Our approach differs from classical conditional GAN \cite{mirza2014conditional} by the fact that the objective of our discriminator is to reconstruct the hidden sensitive attribute from the generator output, whereas the discriminator in classical conditional GAN has to discriminate between the generator output and samples from the true distribution. 

Figure \ref{fig:sanTrain} presents the high-level overview of the training procedure, while Algorithm \ref{GANSanAlgo} describes it in details.
\input{SanitizerTraining}

\input{demo}

The first step corresponds to the training of the sanitizer \san{} (Algorithm \ref{GANSanAlgo}, Lines $7-17$). 
The sanitizer can be seen as the generator similarly to standard GAN but with a different purpose. 
In a nutshell, it learns the empirical distribution of the sensitive attribute and generate a new distribution that concurrently respects two objectives: (1) finding a perturbation that will fool the discriminator in predicting \sensAttr{} while (2) minimizing the damage introduced by the sanitization.
More precisely, the sanitizer takes as input the original dataset \dataset{} (including \sensAttr{} and \decAttr{}) plus some noise $P_z$.
The noise introduced is used to prevent the over-specialization of the sanitizer on the training set while making the reverse mapping of sanitized profiles to their original versions more difficult as the mapping will be probabilistic and not deterministic.
As a result, even if the sanitizer is applied twice on the same profile, it can produced two different modified profiles. 

The second step consists in training the discriminator \dis{} for predicting the sensitive attribute from the data produced by the sanitizer \san (Algorithm \ref{GANSanAlgo}, Lines $18-24$).
The rationale of our approach is that the better the discriminator is at predicting the sensitive attribute \sensAttr, the worse the sanitizer is at hiding it and thus the higher the potential risk of discrimination.

These two steps are run iteratively until convergence of the training.

\paragraph{Training objective of \gansan.} 
Let $\bar{\sensAttr{}}$ be the prediction of \sensAttr{} by the discriminator ($\bar{\sensAttr{}} = D_{isc}(S_{an}(D))$). 
Its objective is to accurately predict \sensAttr{}, thus it aims at minimizing the loss $J^{D_{isc}}(S, \bar{\sensAttr{}}) = d_{disc}(\sensAttr{}, \bar{\sensAttr{}})$.
In practice in our work, we instantiate $d_{disc}$ as the Mean Squared Error (MSE).  

Given an hyperparameter $\alpha$ representing the desired trade-off between the fairness and the fidelity, the sanitizer minimizes a loss combining two objectives:
\begin{align}
    \begin{split}
        &J^{S_{an}}(D, \:S_{an}, \:D_{isc}) =  \\
        &\alpha * d_s(\sensAttr{}, \bar{\sensAttr{}}) + (1 - \alpha) * (d_{r}(D, \:S_{an}(D)))
    \end{split}
\end{align}
in which $d_s$ is $ \frac{1}{2} - BER(D_{isc}(\notSensAttr{}, \decAttr{}), \sensAttrL{})$ on the sensitive attribute. 
The term $\frac{1}{2}$ is due to the objective of maximizing the error of the discriminator (\emph{i.e.}, recall that the optimal value of the BER is $0.5$).

Concerning the reconstruction loss $d_r$, we have first tried the classical Mean Absolute Error (MAE) and MSE losses. 
However, our initial experiments have shown that these losses produce datasets that are highly problematic in the sense that the sanitizer always outputs the same profile whatever the input profile, which protects against attribute inference but renders the profile unusable.  
Therefore, we had to design a slightly more complex loss function. 
More precisely, we chose not to merge the respective losses of these attributes ($e_{\notSensAttr{}_i} = (1-\alpha) * |\notSensAttr{}_i - \bar{\notSensAttr{}_i}|;\quad \bar{A}_i \in \bar{A}, i\in[1,d]$), yielding a vector of attribute losses whose components are iteratively used in the gradient descent.

Hence, each node of the output layer of the generator is optimized to reconstruct a single attribute from the representation obtained from the intermediate layers. 
The vector formulation of the loss is as follows: $\vec{J}^{San} = (e_{A_1}, e_{A_2}, e_{A_3}, ..., e_{A_d}, e_{Y}, \alpha*d_s(\sensAttr{}, \bar{\sensAttr{}}))^{T}$ and the objective is to minimize all its components.

The details of the parameters used for the training are given in Appendices \ref{app:preprocessing} and \ref{app:hyper}.

\subsection{Performance metrics}
\label{subsec:metrics}

The performance of \gansan will be evaluated by taking into account the \emph{fairness enhancement} and the \emph{fidelity} to the original data. 
With respect to fairness, we will quantify it primarily with the inability of a predictor $Adv$, hereafter referred to as the adversary, in inferring the sensitive attribute (\emph{cf.} Section \ref{sec:systemmodel}) using its \textit{Balanced Error Rate (\myBer{})}~\cite{Feldman2014} and its \emph{accuracy \mySac{}} (\emph{cf.}, Section \ref{sec:fairness}).
We will also assess the fairness using metrics (\emph{cf.} Section \ref{sec:systemmodel}) such as \emph{demographic parity} (Equation \ref{eq:demopar}) and \emph{equalized odds} (Equation \ref{eq:eqod}).

To measure the fidelity $\sandist$ between the original and the sanitized data, we have to rely on a notion of distance. 
More precisely, our approach does not require any specific assumption on the distance used, although it is conceivable that it may work better with some than others. 
For the rest of this work, we will instantiate $\sandist$ by the $L_2$-norm as it does not differentiate between attributes.

Note however that a high fidelity is a necessary but not a sufficient condition to imply a good reconstruction of the dataset. 
In fact, as mentioned previously early experiments showed that the sanitizer might find a ``median'' profile to which it will map all input profiles. 
Thus, to quantify the ability of the sanitizer to preserve the diversity of the dataset, we introduce the \emph{\mdist{}} measure, which is defined in the following way :
\begin{align}
    \mdist{}  = \dfrac{\sum_{i=1}^{N}\sum_{j=1}^{N}\sqrt{\sum_{k=1}^{d} (\hat{r}_{i,k} - \hat{r}_{j,k})^2}}{N \times (N-1) \times \sqrt{d}}.
    \label{eq:divers}
\end{align}
While $\sandist{}$ quantifies how different the original and the sanitized datasets are, the diversity measures how diverse the profiles are in each dataset.
We will also provide a qualitative discussion of the amount of damage for a given fidelity and fairness to provide a better understanding of the qualitative meaning of the fidelity.

Finally, we evaluate the loss of utility induced by the sanitization by relying on the accuracy $\myAcc{}$ of prediction on a classification task.
More precisely, the difference in $\myAcc{}$ between a classifier trained on the original data and one trained on the sanitized data can be used as a measure of the loss of utility introduced by the sanitization with respect to the classification task.

%% file: SanitizerTraining.tex
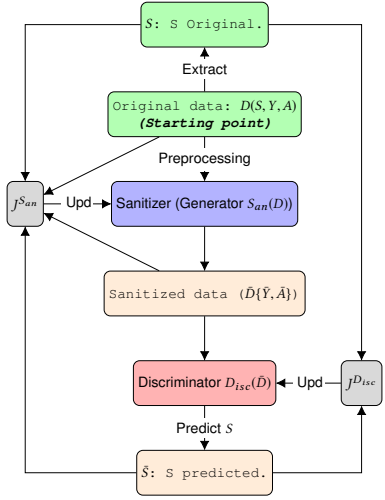
\begin{figure}[h!]
    \centering
    \resizebox{.6\linewidth}{!}{
        \begin{tikzpicture}[node distance=2cm, every node/.style={fill=white, font=\sffamily}, align=center]
            \node (sOrig)            [originals] {$S$: S Original.};
            \node (oSet)             [originals, below of=sOrig] {Original data: $D(S,Y,A)$ \\\textbf{\textit{(Starting point)}}};
            \node (sanitizer)        [sanitizerNode, below of=oSet] {Sanitizer (Generator $S_{an}(D)$)};
            \node (oGenerated)       [process, below of=sanitizer] {Sanitized data ($\bar{D}\{\bar{Y},\bar{A}\}$)};
            \node (disc)             [discriminatorNode, below of=oGenerated] {Discriminator $D_{isc}(\bar{D})$};
            \node (sPred)            [process, below of=disc] {$\bar{S}$: S predicted.};

            \node (jD)               [costs, right of=disc, xshift=1.5cm] {$J^{D_{isc}}$};
            \node (jS)               [costs, left of=sanitizer, xshift=-2cm] {$J^{S_{an}}$};
            
            \draw[->]              (oSet) -- node[text width=4cm]
                                           {Extract}
                                           (sOrig);
            \draw[->]              (oSet) -- node[text width=4cm]
                                           {Preprocessing}
                                           (sanitizer);
            \draw[->]              (sanitizer) -- (oGenerated);
            \draw[->]              (oGenerated) -- (disc);
            \draw[->]              (disc) -- node[text width=2cm]
                                           {Predict $S$}
                                           (sPred);
            
            \draw[->]              (sPred) -| (jD);
            \draw[->]              (sPred) -| (jS);
            \draw[->]              (sOrig) -| (jD);
            \draw[->]              (sOrig) -| (jS);
            
            \draw[->]              (jD) -- node[text width=0.5cm]
                                           {Upd}
                                           (disc);
            \draw[->]              (jS) --  node[text width=0.5cm]
                                           {Upd}
                                           (sanitizer);
            
            \draw[->]              (oSet) -- (jS);
            \draw[->]              (oGenerated) -- (jS);
        \end{tikzpicture}
        }
    \caption{Overview of the framework of \gansan. 
    The objective of the discriminator is to predict \sensAttr\, from the output of the sanitizer $\bar{D}$. 
    The two objective functions that the framework aims at minimizing are respectively the discriminator and sanitizer losses, namely $J^{D_{isc}}$ and $J^{S_{an}}$.}
    \label{fig:sanTrain}
\end{figure}


%% file: demo.tex
\begin{algorithm}
\small
\begin{algorithmic}[1]
\State Inputs: $D=\{\notSensAttr{}, \decAttr{}, \sensAttr{}\}$, $\mathsf{MaxEpochs}$, $d_{iter}$, $\mathsf{batchSize}$, $\alpha$
\State Output: \san{}, \dis{}
\Statex   \Comment {Initialization}
\State \san{}, \dis{}, $Data_{isc} = \mathsf{shuffle}(D)$
\State $\mathsf{Iterations}=\frac{|D|}{\mathsf{batchSize}}$
\For{e $\in$ $\{1,..,\mathsf{MaxEpochs}\}$}
    \For{i $\in$ $\{1,..,\mathsf{Iterations}\}$}
        \State Sample batch B of size $\mathsf{batchSize}$ from $D$
        \State $S_B$: extract S column from $B$
        \State $\{\bar{\notSensAttr{}}, \bar{\decAttr{}}\} = S_{an}(B)$
        \State $e_{\notSensAttr{}_i} = \dfrac{1}{batchSize} \cdot \sum_{n=1}^{\mathsf{batchSize}} |\notSensAttr{}_i^n - \bar{\notSensAttr{}}_i^n|$
        \Statex \Comment {Compute the reconstruction loss vector}
        \State $\vec{J}^{San} = (1-\alpha) \cdot (e_{A_1}, e_{A_2}, e_{A_3}, ..., e_{A_d}, e_{Y})^{T}$
        \Statex \Comment {compute the sensitive loss}
        \State $e_{\sensAttr{}} =  \dfrac{1}{2} - \mathsf{BER}(D_{isc}(S_{an}(B), S_B)$
        \Statex \Comment {concatenate the previously computed loss}
        \State $\vec{J}^{San} = \mathsf{concat}(\vec{J}^{San}, e_{\sensAttr{}})$
    
        \For{$loss$ $\in$ $\vec{J}^{San}$}
            \Statex \Comment{Back-propagation using $loss$}
            \State Backpropagate $loss$ 
            \State Update $S_{an}$ weights
        \EndFor
        \For{l $\in$ $\{1,..,d_{iter}\}$}
            \State Sample batch $B$ of size $batchSize$ from $Data_{isc}$
            \State $S_B$: extract S column from $B$
            \State $Loss$ = MSE($S_B$, $D_{isc}(S_{an}(B))$)
            \State Backpropagate $Loss$
            \State Update $D_{isc}$ weights
        \EndFor
    \EndFor
    \State Save $S_{an}$ and $D_{isc}$ states
\EndFor
\end{algorithmic}
\caption{GANSan Training Procedure}
\label{GANSanAlgo}
\end{algorithm}

%% file: 08-Experiments.tex
\section{Experimental evaluation} 
\label{sec:experiments}

In this section, we describe the experimental setting used to evaluate \gansan as well as the results obtained.

\input{Datasets-Prep}
\input{EvaluationMetrics}
\input{Results}

%% file: Datasets-Prep.tex
\subsection{Experimental setting}

\paragraph{Dataset description.}
We have evaluated our approach on two datasets that are classical in the fairness litterature, namely the \emph{Adult Census Income} as well as on \emph{German Credit}.
Both are available on the UCI repository\footnote{\url{https://archive.ics.uci.edu/ml/index.php}}.
Adult Census reports the financial situation of individuals, with 45222 records after the removal of rows with empty values. 
Each record is characterized by 15 attributes among which we selected the \emph{gender} (\emph{i.e.}, male or female) as the sensitive one and the \emph{income level} (\emph{i.e.}, over or below 50K\$) as the decision.
German Credit is composed of 1000 applicants to a credit loan, described by 21 of their banking characteristics.
Previous work \cite{Kamiran2009} have found that using the \textit{age} as the sensitive attribute by binarizing it with a threshold of $25$ years to differentiate between old and young yields the maximum discrimination based on $\myDI$. 
In this dataset, the decision attribute is the quality of the customer with respect to his credit score (\emph{i.e.}, good or bad).
Due to lack of space, we will mostly discuss the results on Adult dataset in this section. 
However, the results obtained on German credit were quite similar. 

\begin{table*}[h!]
    \centering
    \resizebox{1\linewidth}{!}{
\begin{tabular}{|c|c|c|c|c|}
\cline{1-5}
Dataset
\multirow{2}{*}{}                         & \multicolumn{2}{c|}{\textbf{Adult Census}}                     & \multicolumn{2}{c|}{\textbf{German Credit}}             \\ \cline{1-5} 
                             Group             & Protected ($S_x=S_0$, Female) & Default ($S_x=S_1$, Male) & Protected ($S_x=S_0$, Young) & Default ($S_x=S_1$, Old) \\ \hline
\multicolumn{1}{|c|}{$Pr(S = S_x)$}       & $36.21\%$                   & $63.79\%$                 & $19\%$                      & $81\%$                    \\ \hline
\multicolumn{1}{|c|}{$Pr(Y=1 | S = S_x)$} & $11.35\%$                   & $31.24\%$                 & $57.89\%$                   & $72.83\%$                 \\ \hline
\multicolumn{1}{|c|}{$Pr(Y=1)$}           & \multicolumn{2}{c|}{$24.78\%$}                          & \multicolumn{2}{c|}{$70\%$}                             \\ \hline
\end{tabular}
}
\caption{Distribution of the different groups with respect to the protected attribute and the decision one for both the Adult Census Income and the German Credit datasets.}
\label{table:datastats}
\end{table*}

%% file: EvaluationMetrics.tex
\paragraph{Training process.}
We will evaluate \gansan{} using metrics among which the fidelity $\sandist{}$, the $\myBer{}$ as well as the demographic parity $\myDI{}$ (\emph{cf.} Section \ref{subsec:metrics}). 
For this, we have conducted a $10$-fold cross-validation during which the dataset is divided into ten blocks.
During each fold, 8 blocks are used for the training, while another one is retained as the validation set and the last one as the test set.

We computed the $\myBer{}$ and $\mySac{}$ using the internal discriminator of \gansan{} and three external classifiers independent of the \gansan framework, namely \emph{Support Vector Machines} (SVM)~\cite{Cortes1995}, \emph{Multilayer Perceptron} (MLP)~\cite{PopescuMLP} and \emph{Gradient Boosting} (GB)~\cite{FriedmanGB}. 
For all these external classifiers and all epochs, we report the space of achievable points with respect to the fidelity/fairness trade-off. 
Note that most approaches described in the related work (\emph{cf.} Section \ref{sec:rel_work}) do not validate their results with independent external classifiers trained outside of the sanitization procedure. 

The fact that we rely on three different family of classifiers is not fullproof, in the sense that it might exist another classifiers that we have not tested that can do better, 
but it provides a higher confidence on the strength of the sanitization than simply relying on the internal discriminator.

For each fold and each value of $\alpha$, we train the sanitizer during $40$ epochs. 
At the end of each epoch, we save the state of the sanitizer and generate a sanitized dataset on which we compute the $\myBer{}$, $\mySac{}$ and $\sandist{}$. 
Afterwards, $\heuristic{}$ is used to select the sanitized dataset that is closest to the ``ideal point'' ($\myBer{} = 0.5, \sandist{} = 1$).
More precisely, $\heuristic{}$ is defined as follows: $Best_{Epoch} = min\{(BER_{min} - \dfrac{1}{2})^2 + \sandist{}_{e}, for\: e \in \{1,\ldots,MaxEpoch\}\}$ with $BER_{min}$ referring to the minimum value of $\myBer{}$ obtained with the external classifiers.
For each value of $\alpha \in [0, 1]$, $\heuristic{}$ selects among the sanitizers saved at the end of each epoch, the one achieving the highest fairness in terms of $\myBer{}$ for the lowest damage.
We will use the three families of external classifiers for computing $\myAcc{}$, $\myDI{}$ and $\eqod{}$. 
We also used the same chosen test set to conduct a detailed analysis of its reconstruction's quality ($\mdist{}$ and quantitative damage on attributes).

\subsection{Evaluation scenarios}

Recall that \gansan{} takes as input the whole original dataset (including the sensitive and the decision attributes) and outputs a sanitized dataset (\emph{without} the sensitive attribute) in the same space as the original one, but from which it is impossible to infer the sensitive attribute.
In this context, the overall performance of \gansan{} can be evaluated by analyzing the reachable space of points characterizing the trade-off between the fidelity $\sandist$ to the original dataset and the fairness enhancement.
More precisely, during our experimental evaluation, we will measure the fidelity between the original and the sanitized data, as well as the $\mdist{}$, both in relation with the $\myBer{}$ and $\mySac{}$, computed on this dataset.

However, in practice, our approach can be used in several situations that differ slightly from one another. 
In the following, we detail four scenarios that we believe as representing most of the possible use cases of \gansan{}.
To ease the understanding, we will use the following notation: the subscript $tr$ (respectively $ts$) will denote the data in the training set (respectively test set). 
For instance, $\{Z\}_{tr}$ in which $Z$ can either be $A$, $Y$, $\bar{A}$ or $\bar{Y}$, represents respectively the attributes of the original training set (not including the sensitive and the decision attributes), the decision in the original training set, the attributes the sanitized training set and the decision attribute in the sanitized training set. 
Table \ref{tab:scenarios} describes the composition of the training and the testings sets for these four scenarios.

\begin{table}[h!]
\centering
\caption{Scenarios envisioned for the evaluation of \gansan. Each set is composed of either the original attributes or their sanitized versions, coupled with either the original or sanitized decision.}
\vspace{-2mm}
\resizebox{\columnwidth}{!}{%
    \begin{tabular}{|c|c|c|c|c|}
        \hline
        \multirow{2}{*}{Scenario} & \multicolumn{2}{c|}{Train set composition} & \multicolumn{2}{c|}{Test set composition} \\
        \cline{2-5} 
        & \notSensAttr{} & \decAttr{} & \notSensAttr{} & \decAttr{} \\
        \hline
        \emph{Baseline} & Original & Original & Original & Original \\
        \emph{Scenario 1} & Sanitized & Sanitized & Sanitized & Sanitized \\
        \emph{Scenario 2} & Sanitized & Original & Sanitized & Original \\
        \emph{Scenario 3} & Sanitized & Sanitized & Original & Original \\
        \emph{Scenario 4} & Original & Original & Sanitized & Original \\
        \hline
    \end{tabular}
}
\label{tab:scenarios}
\vspace{-4mm}
\end{table}

\paragraph{Scenario 1 : complete data debiasing.}
This setting corresponds to the typical use of the sanitized dataset, which is the prediction of a decision attribute through a classifier. 
The decision attribute is also sanitized as we assumed that the original decision holds information about the sensitive attribute.
Here, we quantify the accuracy of prediction of $\{\bar{Y}\}_{ts}$ as well as the discrimination represented by the \emph{demographic parity} (Equation \ref{eq:demopar}) and \emph{equalized odds} (Equation \ref{eq:eqod}). 

\paragraph{Scenario 2 : partial data debiasing.}
In this scenario, similarly to the previous one, the training and the test sets are sanitized with the exception that the sanitized decision in both these datasets $\{\bar{A}, \bar{Y}\}$ is replaced with the original one $\{\bar{A}, Y\}$.
This scenario is generally the one considered in the majority of paper on fairness enhancement~\cite{Zemel2013,CRWAA-ALFR2016,madras2018learning}, the accuracy loss in the prediction of the original decision $\{Y\}_{ts}$ between this classifier and another trained on the original dataset without modifications $\{A\}_{tr}$ is a straightforward way to quantify the utility loss due to the sanitization.

\paragraph{Scenario 3 : building a fair classifier.}
This scenario was considered in~\cite{xu2018fairgan} and is motivated by the fact that the sanitized dataset might introduce some undesired perturbations (\emph{e.g.}, changing the education level from Bachelor to PhD).

Thus, a third party might build a fair classifier but still apply it directly on the unperturbed data to avoid the data sanitization process and the associated risks.
More precisely in this scenario, a fair classifier is obtained by training it on the sanitized dataset $\{\bar{A}\}_{tr}$ to predict the sanitized decision $\{\bar{Y}\}_{tr}$. 
Afterwards, this classifier is tested on the original data ($\{A\}_{ts}$) by measuring its fairness through the demographic parity (Equation \ref{eq:demopar}, Section \ref{sec:systemmodel}).
We also compute the accuracy of the fair classifier with respect to the original decision of the test set $\{Y\}_{ts}$.

\paragraph{Scenario 4 : local sanitization.}
The local sanitization scenario corresponds to local use of the sanitizer by the individual himself. 
For instance, the sanitizer could be used as part of a mobile phone application providing individuals with a mean to remove some sensitive attributes from their profile before disclosing it to an external entity.
In this scenario, we assume the existence of a biased classifier, trained to predict the original decision $\{Y\}_{tr}$ on the original dataset $\{A\}_{tr}$.
The user has no control on this classifier, but he is allowed nonetheless to perform the sanitization locally on his profile before submitting it to the existing classifier similarly to the recruitment scenario discussed in the introduction.
This classifier is applied on the sanitized test set $\{\bar{A}\}_{ts}$ and its accuracy is measured with respect to the original decision $\{Y\}_{ts}$ as well as its fairness quantified by $\myDI{}$.

%% file: Results.tex
\subsection{Experimental results}

\paragraph{General results on Adult.}

\begin{figure*}[h!]
    \includegraphics[width=0.8\linewidth, height=0.35\textheight]{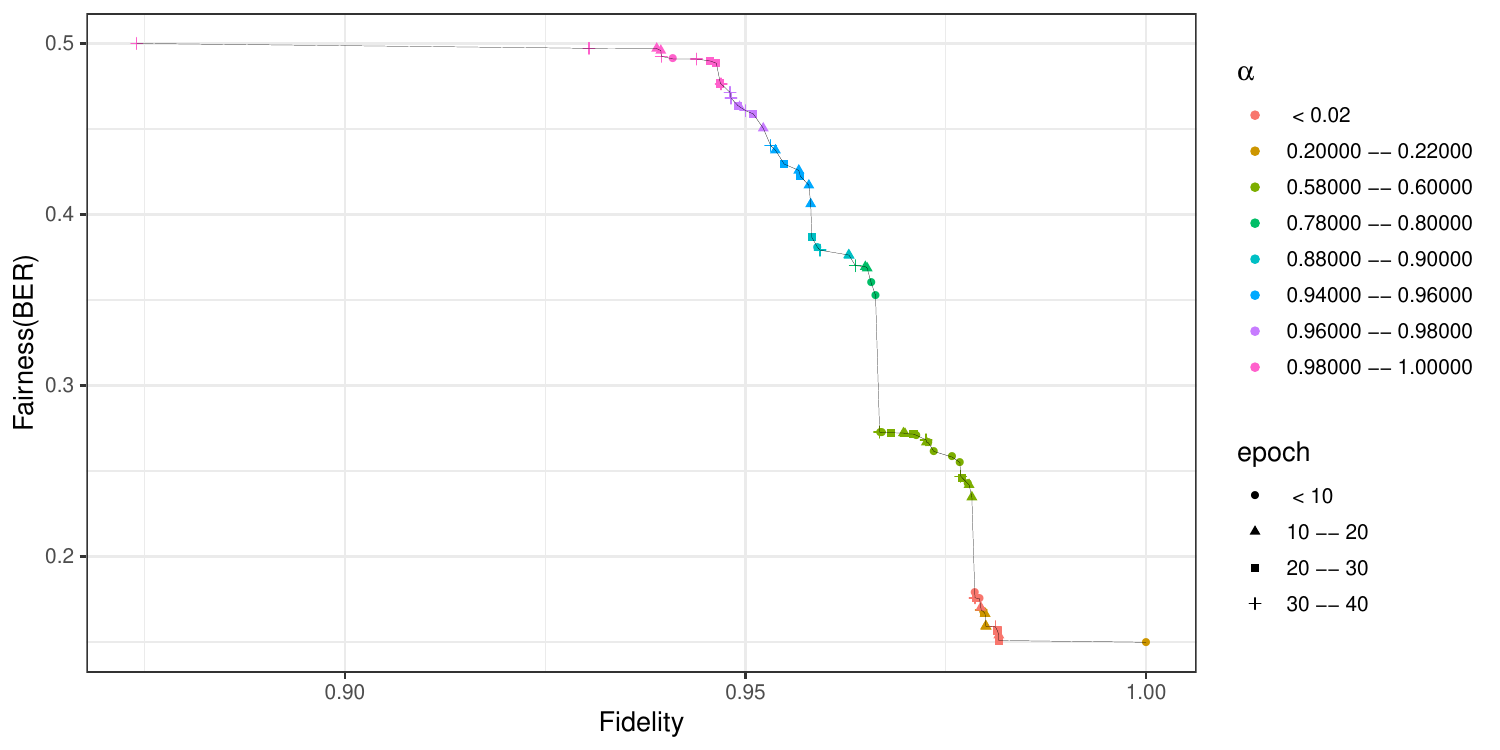}
    \vspace{-2mm}
    \caption{Fidelity/fairness trade-off on Adult. 
Each point represents the minimum possible $\myBer{}$ of all the external classifiers. 
The fairness improves with the increase of $\alpha$, a small value providing a low fairness guarantee while a high one causes greater damage to the sanitized data.}
    \label{res:allValBerLowest}
    \vspace{-4mm}
\end{figure*}

Figure \ref{res:allValBerLowest} describes the achievable trade-off between fairness and fidelity obtained on Adult. 
First, we can observe that fairness improves when $\alpha$ increased as expected.
Even with $\alpha = 0$ (\emph{i.e.}, maximum utility with no focus on the fairness), we cannot reach a perfect fidelity to the original data as we get at most $\sandist{}_{\alpha = 0} \approx 0.982$ (\emph{cf.} Figure \ref{res:allValBerLowest}). 
Increasing the value of $\alpha$ from $0$ to a low value such as $0.2$ provides a fidelity close to the highest possible ($\sandist{}_{\alpha = 0.2} = 0.98$), but leads to a \myBer{} that is poor (\emph{i.e.}, not higher than $0.2$). 
Nonetheless, we still have a fairness enhancement, compared to the original data ($\sandist{}_{orig} = 1$, $\myBer \leq 0.15$).

At the other extreme in which $\alpha = 1$, the data is sanitized without any consideration on the fidelity. 
In this case, the $\myBer{}$ is optimal as expected and the fidelity is $10\%$ lower than the maximum achievable ($\sandist{}_{\alpha = 1} \approx  0.88$).
However, slightly decreasing the value of $\alpha$, such as setting $\alpha = 0.96$, allows the sanitizer to significantly remove the unwarranted correlations ($\myBer{} \approx 0.45$) with a cost of $2.24\%$ on fidelity ($\sandist{}_{\alpha = 0.96} \approx 0.95$).

\begin{figure}[h!]
    \includegraphics[width=1\linewidth, height=0.25\textheight]{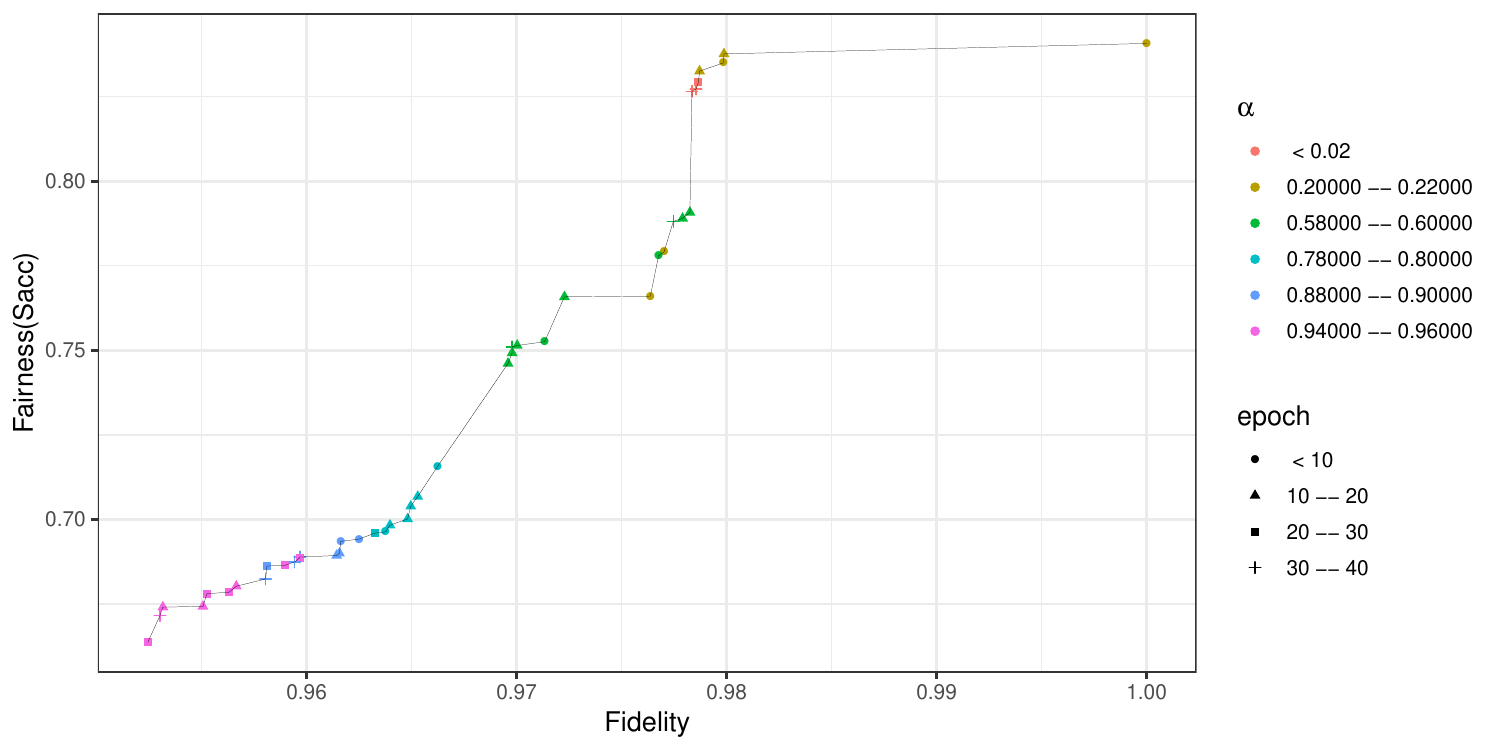}
    \caption{Fidelity-fairness trade-off on Adult. 
    Each point represents the minimum possible $\mySac{}$ of all the external classifiers. 
    $\mySac{}$ decreases with the increase of $\alpha$, a small value providing a low fairness guarantee while a larger one usually introduced a higher damage.
    Remark that even with $\alpha = 0$, a small damage is to be expected. 
Points whose $fidelity=1$ (lower right) represent the $\myBer{}$ on the original (\emph{i.e.}, unperturbed) dataset.}
    \label{res:adultsacc}
\end{figure}

\begin{figure*}[h!]
            \begin{subfigure}[t]{\textwidth}
                \centering
                \includegraphics[width=0.9\linewidth, height=0.35\textheight]{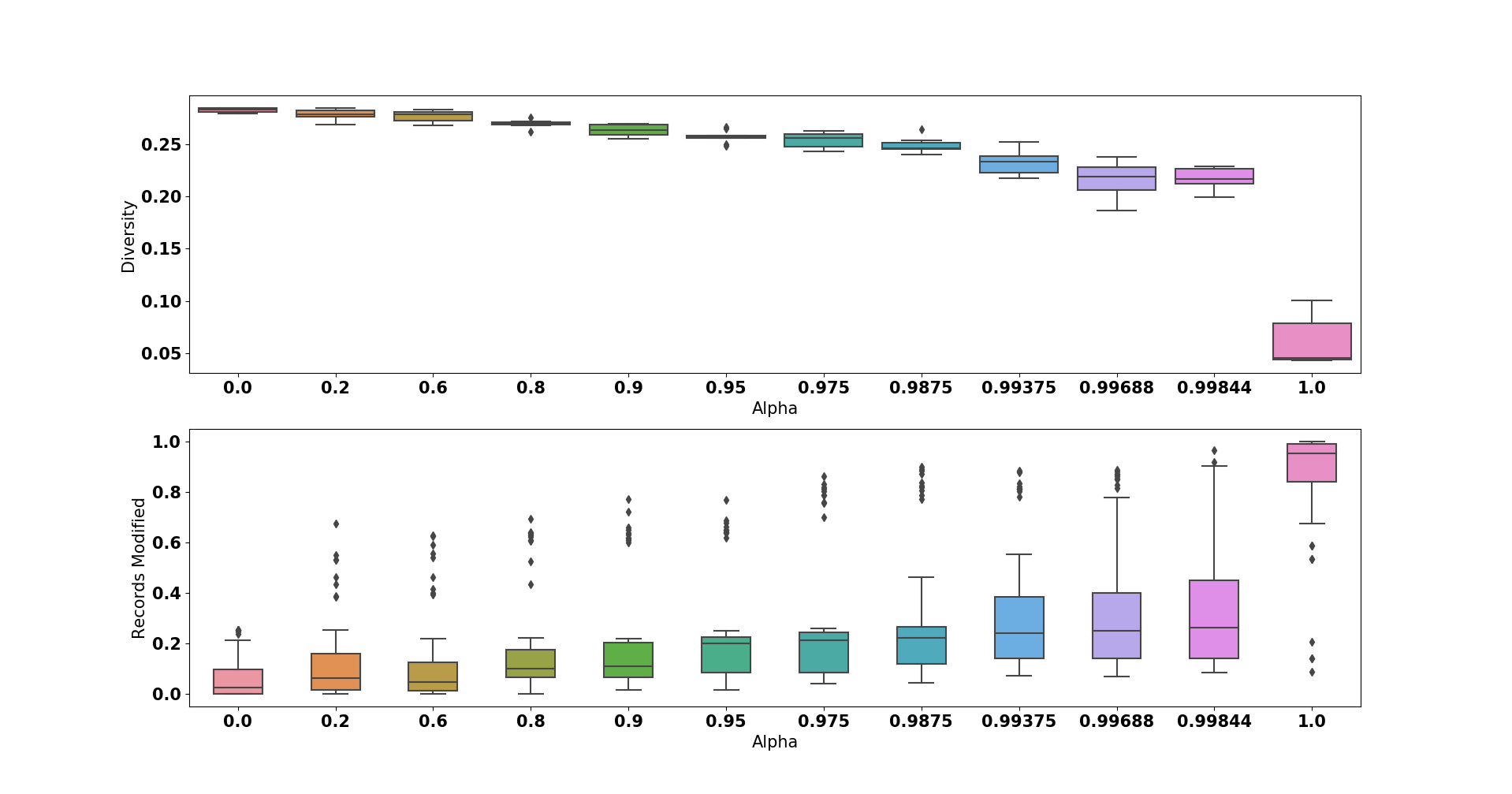}
            \end{subfigure}

        \caption{Boxplots of the quantitative analysis of sanitized datasets selected using $\heuristic{}$. 
        These metrics are computed on the whole sanitized dataset. 
        Modified records correspond to the proportion of records with categorical attributes affected by the sanitization.}
        \label{res:quantAnalysis}
        \vspace{-4mm}
\end{figure*}

With respect to $\mySac{}$, the accuracy drops significantly when the value of $\alpha$ increases \ref{res:adultsacc}.
Here, the optimal value is the proportion of the majority class, which \gansan{} renders the accuracy of predicting \sensAttr{} from the sanitized set closer to that value. 
However, even at the extreme $\alpha = 1$, it is nearly impossible to reach this optimal value. 
Similarly to \myBer{}, slightly decreasing $\alpha$ from this extreme value by setting $\alpha = 0.85$ improves the sanitization while preserving a fidelity closer to the maximum achievable.

The quantitative analysis with respect to the diversity is shown in Figure \ref{res:quantAnalysis}. 
More precisely, the smallest drop of diversity obtained is $3.57\%$, which is achieved when we set $\alpha \leq 0.2$.
Among all values of $\alpha$, the biggest drop observed is $36\%$.
The application of \gansan{}, therefore introduces an irreversible perturbation as observed with the fidelity. 
This loss of diversity implies that the sanitization reinforces the similarity between sanitized profiles as $\alpha$ increases, rendering them almost identical or mapping the input profiles to a small number of stereotypes.
When $\alpha$ is in the range $[0.98, 1[$ (\emph{i.e.}, complete sanitization), $75\%$ of categorical attributes have a proportion of modified records between $10$ and $40\%$ (\emph{cf.} Figure 
\ref{res:quantAnalysis}).

For numerical attributes, we compute the relative change (\textit{RC}) normalized by the mean of the original and sanitized values:
\begin{align}
    RC &= \frac{|original - sanitized|}{f(original, sanitized)} \\
    f(original, sanitized) &= \frac{|original| + |sanitized|}{2}
\end{align}
We normalize the RC using the mean (since all values are positives) as it allows us to handle situations in which the original values are equal to $0$.
With the exception of the extreme sanitization ($\alpha = 1$), at least $70\%$ of records in the dataset have a relative change lower than $0.25$ for most of the numerical attributes.
Selecting $\alpha = 0.9875 \geq 0.98 $ leads to $80\%$ of records being modified with a relative change less than $0.5$ (c.f. Figure ~\ref{res:adultnumDam} in appendix~\ref{res:adultcomplementary}).

\paragraph{General results on German.}
\begin{figure}[h!]
    \includegraphics[width=1\linewidth, height=0.25\textheight]{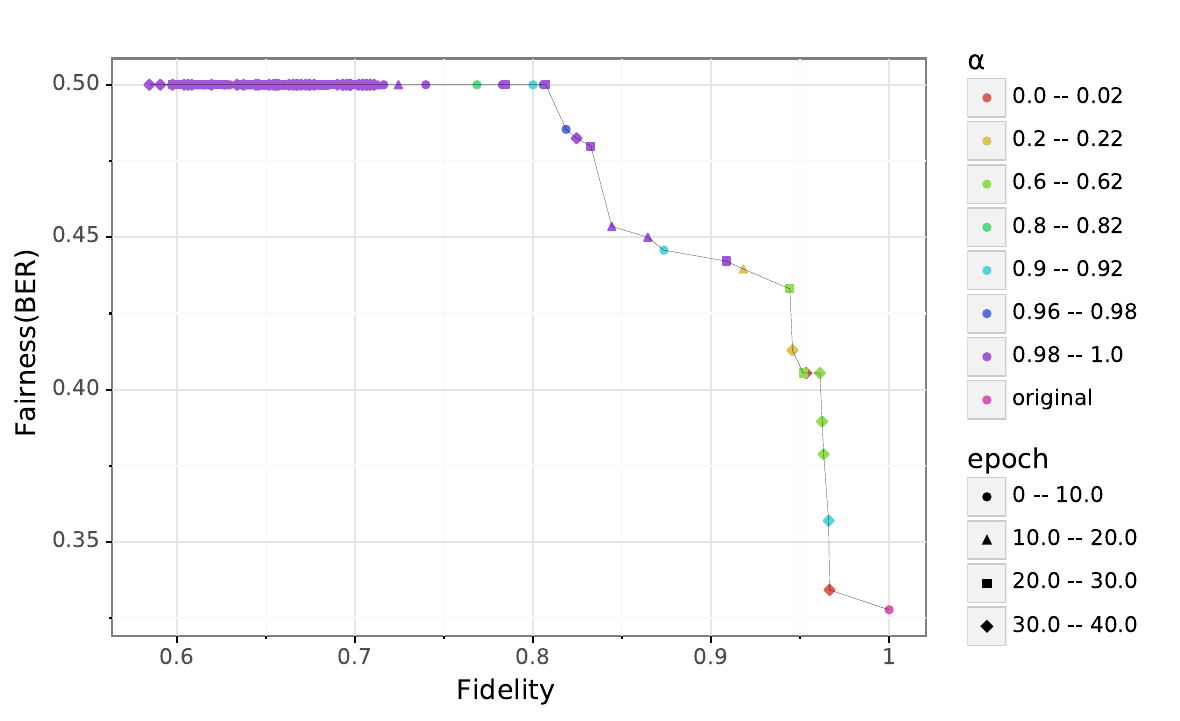}
    \vspace{-8mm}
    \caption{Fidelity/fairness trade-off on German Credit.}
    \label{res:germanber}
\end{figure}

Similarly to Adult, the protection increases with $\alpha$. 
More precisely $\alpha = 0$ (maximum reconstruction) achieves a fidelity of almost $0.96$. 
The maximum protection of $\myBer{} = 0.5$ corresponds to a fidelity of $0.81$ and a sensitive accuracy value of $\mySac{} = 0.76$. 
\begin{figure}[h!]
    \includegraphics[width=1\linewidth, height=0.25\textheight]{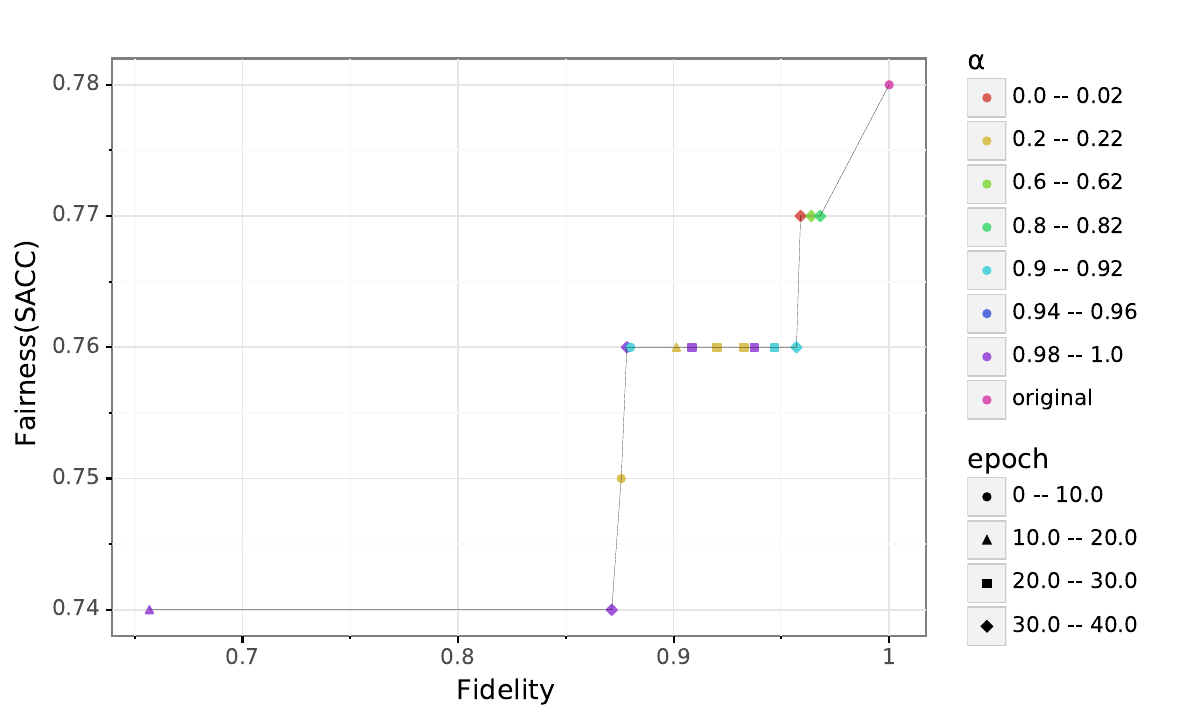}
    \vspace{-8mm}
    \caption{Fidelity-Fairness trade-off on German Credit. 
    Each point represents the minimum possible $\mySac{}$ of all the external classifiers.}
    \label{res:germansac}
\end{figure}

We can observe on Figure\ref{res:germansac} that most values are concentrated on the $0.76$ plateau, regardless of the fidelity and the value of $\alpha$. 
We believe this is due to the high disparity of the dataset.
The fairness on German credit is initially quite high, being close to $0.33$. 
Nonetheless, we can observe three interesting trade-offs on Figure\ref{res:germanber}, each located at a different shoulder of the Pareto front. 
These trade-offs are \textit{A} ($\myBer{} \approx 0.43, \sandist{} \approx 0.94$), \textit{B} ($\myBer{} \approx 0.45, \sandist{} \approx 0.84$) and \textit{C} ($\myBer{} \approx 0.5, \sandist{} \approx 0.81$), each achievable with $\alpha = 0.6$ for the first one, and $\alpha = 0.9968$ for the rest. 

\begin{figure*}[h!]
    \includegraphics[width=0.8\linewidth, height=0.45\textheight]{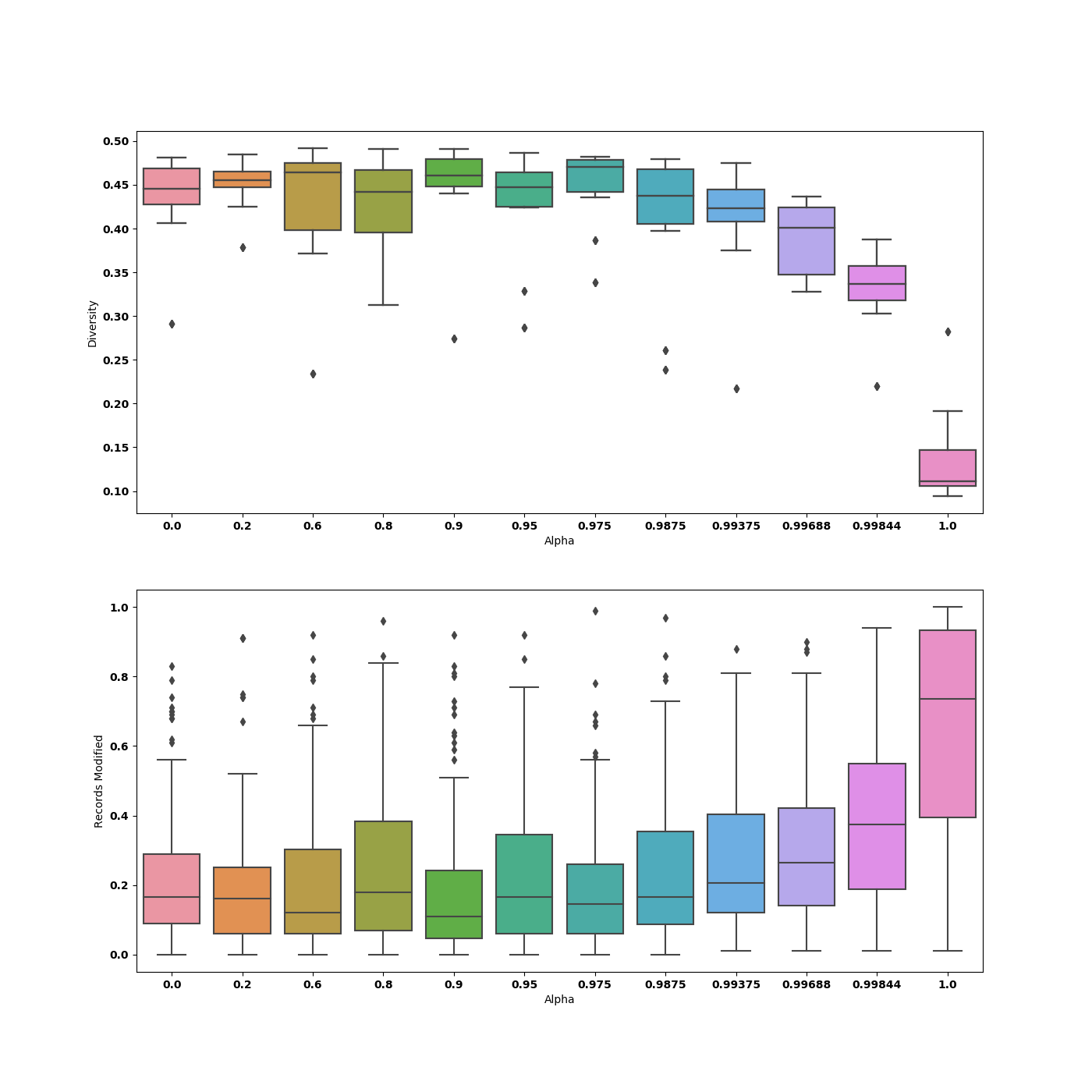}
    \vspace{-8mm}
    \caption{Diversity and categorical damage on German.}
    \label{res:germandiv}
\end{figure*}

We review the diversity and the sanitization induced damage on categorical attributes in Figure~\ref{res:germandiv}. 
As expected, the diversity decreases with alpha, rendering most profiles identical with $\alpha = 1$.
We can also observe some instabilities  higher $\alpha$ values produce a shallow range of diversities (i.e $\alpha \geq 0.9$) while smaller values have a higher range of diversities. 
Such instability is mainly explained by the size and the imbalance of the dataset, which does not allow the sanitizer to correctly learn the distribution (such phenomenon is common when training GANs with a small dataset). 
Nonetheless, most of the diversity results  prove close to the original one, that is $0.51$. 
The same trend is observed on the categorical attribute damage. For most values of $\alpha$, the median damage is below or equal to $20\%$, meaning that we have to modify only two categorical columns in a record to prevent remove unwanted correlations. 
For the numerical damage, most columns have a relative change lower than $0.5$ for more than $70\%$ of the dataset, regardless of the value of $\alpha$. Only columns \textit{Duration in month} and \textit{Credit amount} have a higher damage. 
This is due to the fact that these columns have a very large range of possible values compare to the other columns ($33$ and $921$), especially for column \textit{Credit amount} which also exhibit a nearly uniform distribution.
Our reference points \textit{A}, \textit{B} and \textit{C} have a median damage close than $10\%$ for \textit{A} and $20\%$ for both \textit{B} and \textit{C}. 
The damage on categorical columns are also acceptable.

To summarize our results, \gansan{} is able to maintain an important part of the dataset structure despite sanitization, making it usable for other analysis tasks. 
However, at the individual level, some perturbation might be more important on some profiles than on others. 
A future work will investigate the relationship between the position of the profile in distributions and the damage introduced.
For the different scenarios investigated hereafter, we fixed the value of $\alpha$ to $0.9875$, which provides nearly a perfect level of sensitive attribute protection while leading to an acceptable damage on Adult. 
Due to space limitations, we will not discussed results obtained on German, the scenario analysis are available on the appendices~\ref{app:scenariofairgerman}.

\begin{figure*}[h!]
    \centering
    \includegraphics[width=\linewidth]{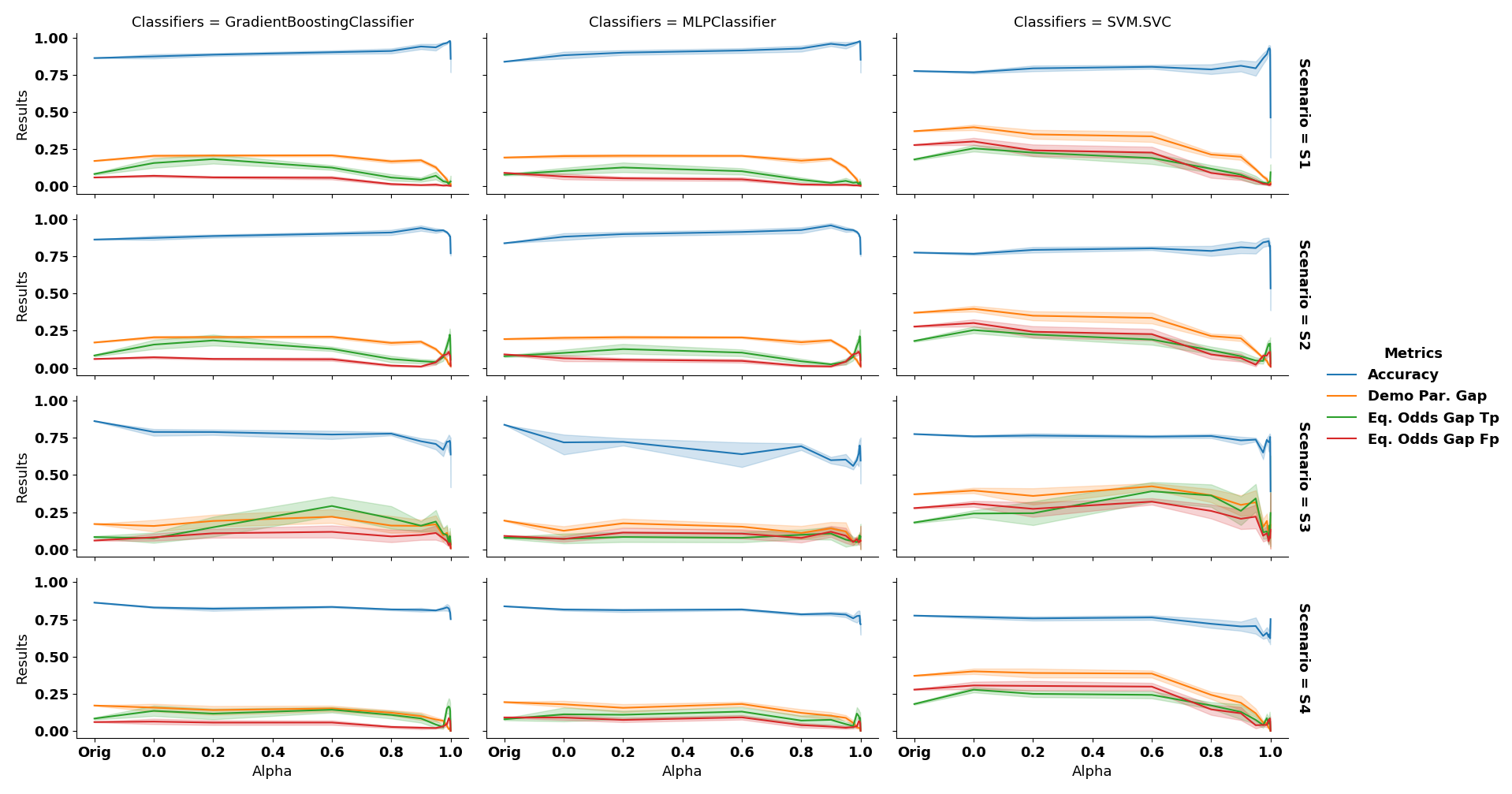}

    \caption{Accuracy (blue), demographic parity gap (orange) and equalized odds gap (true positive rate in green and false positive rate in red) computed for scenarios 1, 2, 3 and 4 (top to bottom), with the classifiers GB, MLP and SVM (left to right) on Adult dataset. 
    The greater the value of $\alpha$ the better the fairness. Using only the sanitized data $\bar{\notSensAttr{}}$ (\textit{S1, S2}) increases the accuracy while a combination of the original ($\notSensAttr{}$) and sanitized data ($\bar{\notSensAttr{}}$) decreases it.}
    \label{res:all_scn}
    \vspace{-4mm}
\end{figure*}

\paragraph{Scenario 1 : complete data debiasing.}
In this scenario, we observe that \gansan{} preserves the accuracy of the dataset.
More precisely, it increases the accuracy of the decision prediction on the sanitized dataset for all classifiers (\emph{cf.} Figure \ref{res:all_scn}, Scenario S1), compared to the original one which is $0.86$, $0.84$ and $0.78$ respectively for GB, MLP and SVM. 
This increase can be explained by the fact that \gansan{} modifies the profiles to make them more coherent with the associated decision, by removing correlations between the sensitive attribute and the decision one.
As a consequence, this sets the same decision to similar profiles in both the protected and the default groups. 
In fact, nearly the same distributions of decision attribute are observed before and after the sanitization but some record's decisions are shifted ($7.56\% \pm 1.23\%$ of decision shifted in the sanitized whole set, $11.44\% \pm 2.74\%$ of decision shifted in the sanitized sensitive group for $\alpha = 0.9875$). 
Such decision shift could be explained by the similarity between those profiles to others with the opposite decisions in the original dataset.

We also believe that the increase of accuracy is correlated with the drop of diversity. 
More precisely, if profiles become similar to each other, the decision boundary might be easier to find. 

The discrimination is reduced as observed through $\myDI{}$, $\eqod_1$ and $\eqod_0$, which all exhibit a negative slope. 
When correlations with the sensitive attribute are significantly removed ($\alpha \geqslant 0.6$), those metrics also significantly decrease. 
For instance, at $\alpha=0.9875$, $\myBer{} \geq 0.48$, $\myAcc{} = 0.96$, $\myDI{} = 0.0453$, $\eqod_1 = 0.0286 $ $\eqod_0 = 0.0062$ for GB; whereas as the original demographic parity gap and equalised odds gap are respectively $\myDI{} = 0.16$, $\eqod_1 = 0.083$ $\eqod_0 = 0.060$ (\emph{cf.}, Tables \ref{tab:eqod} and \ref{tab:protection_level} in appendices for more details).
In this setup, \textsf{FairGan} \cite{xu2018fairgan} achieves a \myBer{} of $0.3862 \pm 0036$ an accuracy of $0.82 \pm 0.01$ and a demographic parity of $0.04 \pm 0.02$.

\paragraph{Scenario 2 : partial data debiasing.}

Somewhat surprisingly, we observe an increase in accuracy for most values of alpha. 
The demographic parity also decreases while the equalized odds remains nearly constant ($\eqod_1$, green line on Figure \ref{res:all_scn}). 
Table \ref{tab:demo_par_soa} compare the results obtained to other existing work from the state-of-the-art. 
We include the classifier with the highest accuracy (MLP) and the one with the lowest one (SVM). 
From these results, we can observe that our method outperforms the others in terms of accuracy, but that the lowest demographic parity is achieved with the method proposed in~\cite{zhang2018mitigating} ($\myDI{} = 0.01$), which is not surprising as this method is precisely tailored to reduce this metric.

Even though our method is not specifically constrained to mitigate the demographic parity, we can observe that it significantly improve it. 
Thus, while partial data debiasing is not the best application scenario for our approach as the original decision might be correlated with the sensitive attribute, it still mitigates its effect to some extent.

\begin{table}[h!]
\centering
\caption{Comparison with other works on the basis of accuracy and demographic parity on Adult.}
\vspace{-2mm}
\resizebox{\columnwidth}{!}{%
    \begin{tabular}{|c|c|c|}
    \hline
    \textit{Authors} & \textit{\myAcc{}} & \textit{\myDI{}} \\\cline{1-3}
    \textit{LFR} ~\cite{Zemel2013} & $0.78$ & $\approx 0.02$\\\cline{1-3}
    \textit{ALFR} ~\cite{CRWAA-ALFR2016} & $0.825$ & $\approx 0.02$\\\cline{1-3}
    \textit{MUBAL}~\cite{zhang2018mitigating} & $0.82$ & \textbf{0.01}\\\cline{1-3}
    \textit{LATR}~\cite{madras2018learning} & \textbf{0.84} & $0.10$\\\cline{1-3}
    \gansan{} \textit{(S2) - MLP, $\alpha = 0.9875$} & \textbf{0.91 $\pm$ 0.01} & $0.050 \pm 0.02$\\\cline{1-3}
    \gansan{} \textit{(S2) - SVM, $\alpha = 0.9875$} & \textbf{0.85 $\pm$ 0.04} & $0.048 \pm 0.02$\\\cline{1-3}
    \end{tabular}
}
\label{tab:demo_par_soa}
\vspace{-5mm}
\end{table}

\paragraph{Scenario 3 : building a fair classifier.}
The sanitizer helps to reduce discrimination based on the sensitive attribute, even when using the original data on a classifier trained on the sanitized one. 
As presented in the third row of Figure \ref{res:all_scn}, as we force the system to completely remove the unwarranted correlations, the discrimination observed when classifying the original unperturbed data is reduced.
On the other hand, the accuracy exhibits here the highest negative slope with respect to all the scenarios investigated. 
More precisely, we observe a drop of $16\%$ for the best classifier in terms of accuracy on the original set, which can be explained by the difference of correlations between $A$ and $Y$ and between $\bar{A}$ and $\bar{Y}$. 
As the fair classifiers are trained on the sanitized set ($\bar{A}$ and $\bar{Y}$), the decision boundary obtained is not relevant for $A$ and $Y$. 

\textsf{FairGan} \cite{xu2018fairgan}, which also investigated this scenario achieved $\myAcc{} = 0.82$ and $\myDI{} = 0.05 \pm 0.01$ whereas our best classifier in accuracy (GB) achieves $\myAcc{} = 0.72 \pm 0.033$ and $ \myDI{} = 0.12 \pm 0.06$ for $\alpha = 0.9875$.

\paragraph{Scenario 4 : local sanitization.}
On this setup, we observe that the discrimination is lowered as the $\alpha$ coefficient increases. 
Similarly to other scenarios, the more the correlations with the sensitive attribute are removed, the higher the drop of discrimination as quantified by the $\myDI{}$, $\eqod_1$ as well as $\eqod_0$, and the lower the accuracy on the original decision attribute. 
For instance, with GB, we obtain $\myAcc{} = 0.83 \pm 0.039$, $\myDI{} = 0.035 \pm 0.022$ at $\alpha = 0.9875$ (the original values were $\myAcc{} = 0.86$ and $\myDI{} = 0.16$). 
With MLP which has the best \myDI{}, we observe: $\myAcc{} = 0.77 \pm 0.060$, $\myDI{} = 0.025 \pm 0.017$
This proves that \gansan{} can be used locally, thus allowing users to contribute to large datasets by sanitizing and sharing their information without relying on any third party, with the guarantee that the sensitive attribute \gansan{} has been trained for is removed. 

The drop of accuracy due to the local sanitization is $3.68\%$ on GB ($8\%$ with MLP). 
Thus, for application requiring a time-consuming training phase, using \gansan{} to sanitize profiles without retraining the classifier seems to be a good compromise.

%% file: 09-Conclusion-FutureWork.tex
\section{Conclusion} 
\label{discuss_conclude}
In this work, we have introduced \gansan, a novel sanitization method inspired by GANs achieving fairness by removing the correlations between the sensitive attribute and the other attributes of the profile.
Our experiments demonstrate that \gansan{} can prevent the inference of the sensitive attribute while limiting the loss of utility as measured in terms of the accuracy of a classifier learned on the sanitized data as well as by the damage on the numerical and categorical attributes. 
In addition, one of the strengths of our approach is that it offers the possibility of local sanitization, by only modifying the attributes as little as possible while preserving the space of the original data (thus preserving interpretability).
As a consequence, \gansan{} is agnostic to subsequent use of data as the sanitized data is not tied to a particular task. 

While we have relied on three different types of external classifiers for capturing the difficulty to infer the sensitive attribute from the sanitized data, it is still possible that a more powerful classifier exists that could infer the sensitive attribute with higher accuracy. 
Note that this is an inherent limitation of all the preprocessing techniques and not only our approach.
Nonetheless, as future work we would like to investigate other families of learning algorithms to complete the range of external classifiers.

Finally, much work still needs to be done to assess the relationship between the different fairness notions, namely the impossibility of inference and the individual and group fairness. 

%% file: myBib.tex
\bibliographystyle{ACM-Reference-Format}
\bibliography{library}

%% file: 10-Appendices.tex
\appendix

\section{Preprocessing of datasets}
\label{app:preprocessing}
The preprocessing step consists in first in one-hot encoding categorical and numerical attributes with less than 5 values followed with a scaling between $0$ and $1$. 

In addition on Adult dataset, we need to apply a logarithm on columns $capital-gain$ and $capital-loss$ prior any step because of the fact that those attributes exhibit a distribution close to a Dirac delta~\cite{Dirac1930-DIRTPO}, with the maximal values being respectively $9999$ and $4356$, and a median of $0$ for both (respectively $91\%$ and $95\%$ of records have a value of $0$). 
Since most values are equal to $0$, the sanitizer will always nullify both attributes and the approach will not converge. 
Afterwards, a postprocessing step consisting of reversing the preprocessing ones is performed in order to remap the generated data to the original shape.

\section{Hyper-parameters tuning}
\label{app:hyper}

Table \ref{table:hyperparamsbestadult} details the parameters of the classifiers that have yielded the best results respectively on the Adult and German credit datasets. 
The training rate represents the number of time for which an instance is trained during a single iteration. For instance,  for an iteration $i$, the discriminator is trained with $100*50 = 5000$ records while the sanitizer is trained with $1*100=100$ records. 
The number of iterations is equal to: $iterations = \dfrac{dataset size}{batch size}$.
Our experiments were run for a total of 40 epochs. We varied the $\alpha$ value using a geometric progression: $\alpha_i = 0.2 + 0.4\dfrac{2^i - 1}{2^{i - 1}};$ $i \in \{1,..,10\}$
\begin{table}[h!]
    \centering
    \caption{Hyper parameters tuning for Adult dataset.}
    \resizebox{0.9\columnwidth}{!}{%
    \begin{tabular}{ |c|c|c| } 
        \hline
           & Sanitizer & Discriminator \\ [0.5ex] 
        \hline
            Layers & 3x Linear & 5x Linear \\
            Learning Rate (LR) & $2e-4$ & $2e-4$\\
            Hidden Activation & ReLU & ReLU \\
            Output Activation & LeakyReLU & LeakyReLU\\
            Losses & VectorLoss & MSE \\
            Training rates & 1 & 50 \\
            Batch size & $100$ & $100$ \\
            Optimizers & Adam & Adam \\
        \hline
    \end{tabular}
    }
    \label{table:hyperparamsbestadult}
\end{table}

\section{Evaluation of Adult}
\label{res:adultcomplementary}
This appendix is composed of supplementary results of the evaluation of the Adult dataset.

\subsection{Numerical attribute Damage}

Figure~\ref{res:adultnumDam} summarizes the numerical damage on Adult computed with the formula detailed in Appendix \ref{app:relC}.

\begin{figure*}[h!]
    \includegraphics[width=0.85\linewidth, height=0.35\textheight]{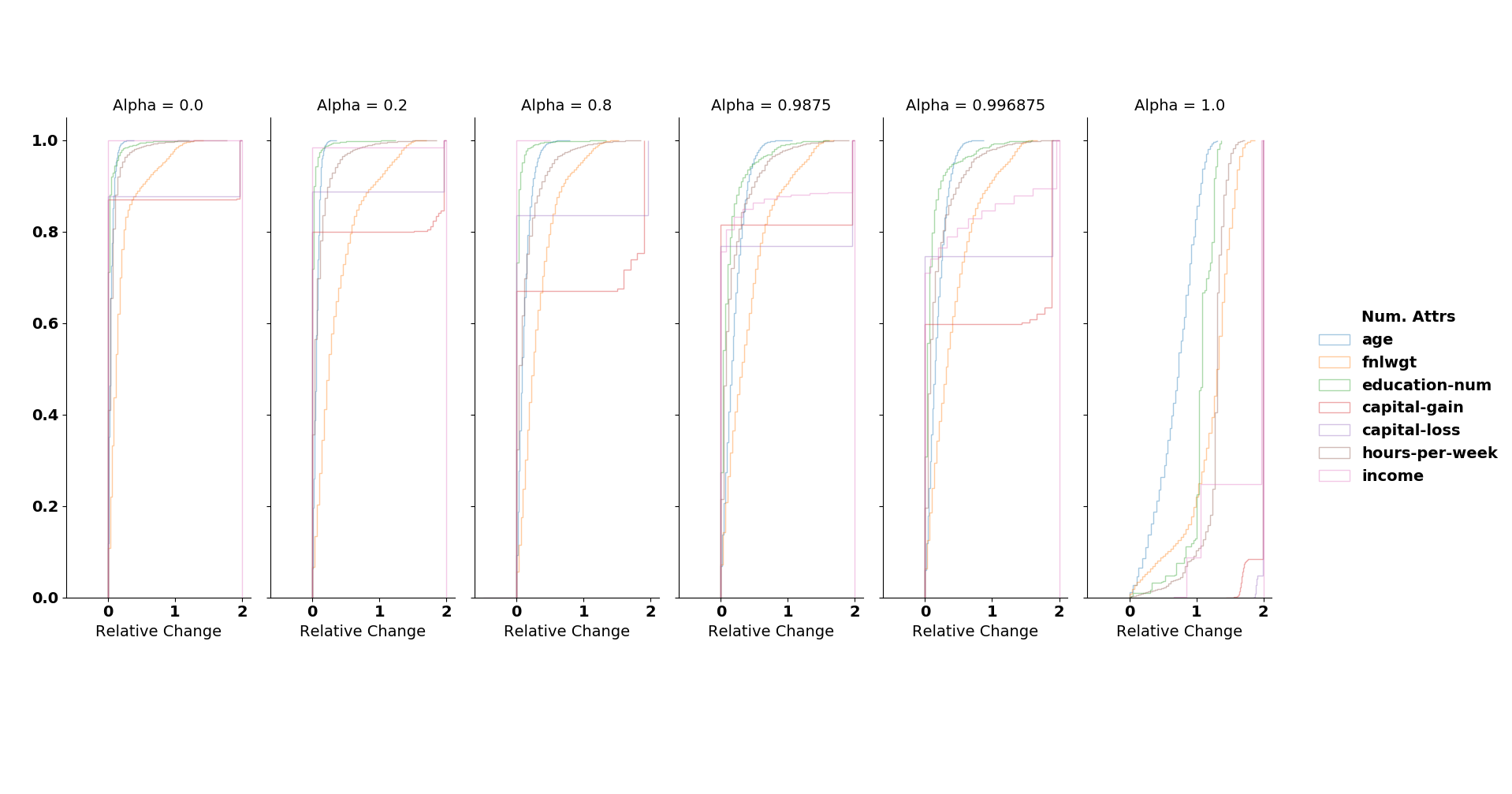}
    \vspace{-12mm}
    \caption{Cumulative distribution of the relative change (x-axis) for numerical attributes, versus the  proportion of records affected in the dataset (y-axis).}
    \label{res:adultnumDam}
\end{figure*}

\subsection{Evaluation of group-based discrimination}
\label{app:results_disc}

Table \ref{tab:eqod} summarizes the results obtained in terms of discrimination, and table \ref{tab:protection_level} presents the sensitive attribute level for all classifiers. These results are computed with $\alpha = 0.9875$.

\begin{table}[h!]
\caption{Equalized odds and demographic parity on Adult.}
\vspace{-4mm}
\centering
\resizebox{\linewidth}{!}{%
    \begin{tabular}{|c|c|c|c|c|c|}
        \hline
        \multirow{2}{*}{Clfs.} & \multicolumn{5}{c|}{$\eqod_1$}\\ \cline{2-6} 
            & \textit{Baseline} & \textit{S1} & \textit{S2} & \textit{S3} & \textit{S4}\\ \hline
        GB  & $0.0830 \pm 0.0374$ & $0.0286 \pm 0.0253$ & $0.1466 \pm 0.0647$ & $0.0966 \pm 0.1044$ & $0.1509 \pm 0.0578$ \\ \cline{2-6}
        SVM & $0.1809 \pm 0.0323$ & $0.0195 \pm 0.0198$ & $0.1249 \pm 0.0668$ & $0.1208 \pm 0.0754$ & $0.0854 \pm 0.0525$ \\ \cline{2-6}
        MLP & $0.0782 \pm 0.0356$ & $0.0266 \pm 0.0176$ & $0.1473 \pm 0.0664$ & $0.0487 \pm 0.0383$ & $0.1165 \pm 0.0680$ \\ \cline{2-6}\hline
    \end{tabular}
}
\resizebox{\linewidth}{!}{%
    \begin{tabular}{|c|c|c|c|c|c|}
        \hline
        \multirow{2}{*}{Clfs.} & \multicolumn{5}{c|}{$\eqod_0$}\\ \cline{2-6} 
            & \textit{Baseline} & \textit{S1} & \textit{S2} & \textit{S3} & \textit{S4}\\ \hline
        GB  & $0.0596 \pm 0.0088$ & $0.0062 \pm 0.0050$ & $0.0907 \pm 0.0274$ & $0.0557 \pm 0.0482$ & $0.0461 \pm 0.0266$ \\ \cline{2-6}
        SVM & $0.2778 \pm 0.0174$ & $0.0149 \pm 0.0113$ & $0.0830 \pm 0.0293$ & $0.1083 \pm 0.0846$ & $0.0402 \pm 0.0273$ \\ \cline{2-6}
        MLP & $0.0905 \pm 0.0155$ & $0.0065 \pm 0.0051$ & $0.0975 \pm 0.0313$ & $0.0695 \pm 0.0274$ & $0.0310 \pm 0.0212$ \\ \cline{2-6}\hline
    \end{tabular}
}
\resizebox{\linewidth}{!}{%
    \begin{tabular}{|c|c|c|c|c|c|}
        \hline
        \multirow{2}{*}{Clfs.} & \multicolumn{5}{c|}{\textit{\myDI}}\\ \cline{2-6} 
           & \textit{Baseline} & \textit{S1} & \textit{S2} & \textit{S3} & \textit{S4}\\ \hline
        GB & $0.1707 \pm 0.0114$ & $0.0453 \pm 0.0261$ & $0.0531 \pm 0.0245$ & $0.1111 \pm 0.0594$ & $0.0352 \pm 0.0224$ \\ \cline{2-6}
        SVM & $0.3709 \pm 0.0139$ & $0.0481 \pm 0.0243$ & $0.0480 \pm 0.0258$ & $0.1910 \pm 0.0845$ & $0.0510 \pm 0.0234$ \\ \cline{2-6}
        MLP & $0.1936 \pm 0.0209$ & $0.0458 \pm 0.0258$ & $0.0508 \pm 0.0253$ & $0.0616 \pm 0.0466$ & $0.0254 \pm 0.0170$ \\ \cline{2-6}\hline
    \end{tabular}
}
\label{tab:eqod}
\end{table}

\begin{table}[h!]
\centering
\caption{Evaluation of \gansan's sensitive attribute protection on Adult.}
\vspace{-4mm}
\resizebox{\linewidth}{!}{%
    \begin{tabular}{|c|c|c|c|c|}
        \hline
        \multirow{2}{*}{Clfs.} & \multicolumn{2}{c|}{\textit{\myBer}} & \multicolumn{2}{c|}{\textit{\mySac}} \\ \cline{2-5} 
            & \textit{Baseline} & \textit{Sanitized} & \textit{Baseline} & \textit{Sanitized}\\ \hline
        GB  & $0.1637 \pm 0.0094$ & $0.4803 \pm 0.0173$ & $0.8530 \pm 0.0074$ & $0.6841 \pm 0.0105$ \\ \cline{2-5}
        MLP & $0.1818 \pm 0.0096$ & $0.4756 \pm 0.0224$ & $0.8423 \pm 0.0034$ & $0.6803 \pm 0.0105$ \\ \cline{2-5}
        SVM & $0.1431 \pm 0.0047$ & $0.4654 \pm 0.0115$ & $0.8255 \pm 0.0052$ & $0.5494 \pm 0.0386$ \\ \cline{2-5}\hline
    \end{tabular}
}
\label{tab:protection_level}
\end{table}

\subsection{Utility of \gansan}
\label{app:results_util}

We present in Table \ref{tab:damage} the utility of \gansan{} as measured in terms of the accuracy on the decision prediction, the fidelity and the diversity on Adult.

\begin{table}[h!]
\centering
\caption{Evaluation of \gansan{}'s utility on adult Census.}
\resizebox{\linewidth}{!}{%
\begin{tabular}{|c|c|c|c|c|c|}
    \hline
    \multirow{2}{*}{Clfs.} & \multicolumn{5}{c|}{\textit{\myAcc}} \\ \cline{2-6} 
        & \textit{Baseline} & \textit{S1} & \textit{S2} &\textit{S3} & \textit{S4} \\ \hline
    GB  & $0.8631 \pm 0.0039$ & $0.9650 \pm 0.0129$ & $0.9119 \pm 0.0116$ & $0.7244 \pm 0.0380$ & $0.8313 \pm 0.0397$ \\ \cline{2-6}
    SVM & $0.7758 \pm 0.0061$ & $0.8895 \pm 0.0502$ & $0.8489 \pm 0.0476$ & $0.7368 \pm 0.0249$ & $0.6605 \pm 0.0649$ \\ \cline{2-6}
    MLP & $0.8384 \pm 0.0030$ & $0.9685 \pm 0.0107$ & $0.9143 \pm 0.0136$ & $0.6008 \pm 0.0464$ & $0.7724 \pm 0.0638$ \\ \hline
\end{tabular}
}
\resizebox{\linewidth}{!}{%
\begin{tabular}{|c|c|c|c|}
    \hline
    \multicolumn{2}{|c|}{\textit{\sandist{}}} & \multicolumn{2}{c|}{\textit{\mdist{}}} \\ \hline
    \textit{Baseline} & \textit{S1} & \textit{Baseline} & \textit{S1} \\ \hline
    $0.852\pm 0.00$ & $0.9428 \pm 0.0025$ & $0.2905$ & $0.2483 \pm 0.0070$ \\ \hline
\end{tabular}
}
\label{tab:damage}
\end{table}

\subsection{Qualitative observation of \gansan{} output on Adult}

In Tables \ref{tab:mostDamPro} and \ref{tab:MinDamMidDamMostDamPro}, we present the records that have been maximally and minimally damaged due to the sanitization.

\begin{table}[h!]
\centering
\caption{Most damaged profiles for $\alpha = 0.9875$ on the first and the fourth folds. 
Only the perturbed attributes are shown.}
\resizebox{\columnwidth}{!}{%
    \begin{tabular}{|c|c|c|c|c|}
    \hline
    \textit{Attrs} & \textit{Original} & \textit{Fold 1} & \textit{Original} & \textit{Fold 4}\\
    \hline \hline 
    {\small age} & {\small 42} & {\small 49.58} & {\small 29} & {\small 49.01} \\
    {\small workclass} & {\small State} & {\small Federal} & {\small Self-emp-not-inc} & {\small Without-pay}\\
    {\small fnlwgt} & {\small 218948} & {\small 192102.77}  & {\small 341672} & {\small 357523.5}\\
    {\small education} & {\small Doctorate} & {\small Bachelors} & {\small HS-grad} & {\small Doctorate}\\
    {\small education-num} & {\small 16} & {\small 9.393} & {\small 9} & {\small 7.674}\\
    {\small marital-status} & {\small Divorced} & {\small Married-civ-spouse} & {\small Married-spouse-absent} & {\small Married-civ-spouse}\\
    {\small occupation} & {\small Prof-specialty} & {\small Adm-Clerical} & {\small Transport-moving} & {\small Protective-serv}\\
    {\small relationship} & {\small Unmarried} & {\small Husband} & {\small Other-relative} & {\small Husband}\\
    {\small race} & {\small Black} & {\small White} & {\small Asian-Pac-Islander} & {\small Black}\\
    {\small hours-per-week} & {\small 36} & {\small 47.04} & {\small 50} & {\small 40.37}\\
    {\small native-country} & {\small Jamaica} & {\small Peru} & {\small India} & {\small Thailand}\\
    {\small damage value} & $-$ & {\small$3.7706$} & {\small India} & {\small Thailand}\\
    \hline
    \end{tabular}
}
\label{tab:mostDamPro}
\end{table}

\begin{table}[h!]
\centering
\caption{Minimally damaged profile, profile with damage at $50\%$ of the max at $\alpha = 0.9875$ for the first fold.}
\resizebox{\columnwidth}{!}{%
    \begin{tabular}{|c|c|c|c|c|}
    \hline
    \textit{Attrs} & \textit{Original} & \textit{$Damage = 0.0291$} & \textit{Original} & \textit{$Damage = 1.8845$} \\
    \hline \hline 
    {\small age} & {\small 49} & {\small 49.4} & {\small 35} & {\small 29.768} \\
    {\small workclass} & {\small Federal-gov} & {\small Federal-gov} & {\small Private} & {\small Private} \\
    {\small fnlwgt} & {\small 157569} & {\small 193388} & {\small 241998} & {\small 179164}\\
    {\small education} & {\small HS-grad} & {\small HS-grad} & {\small HS-grad} & {\small HS-grad}\\
    {\small education-num} & {\small 9} & {\small 9.102} & {\small 9} & {\small 8.2765} \\
    {\small marital-status} & {\small Married-civ-spouse} & {\small Married-civ-spouse} & {\small Never-married} & {\small Never-married} \\
    {\small occupation} & {\small Adm-Clerical} & {\small Adm-Clerical} & {\small Sales} & {\small Farming-fishing} \\
    {\small relationship} & {\small Husband} & {\small Husband} & {\small Not-in-Family} & {\small Not-in-Family} \\
    {\small race} & {\small White} & {\small White} & {\small White} & {\small White} \\
    {\small capital-gain} & {\small 0} & {\small 0} & {\small 8.474} & {\small 0} \\
    {\small capital-loss} & {\small 0} & {\small 0} & {\small 0} & {\small 0} \\
    {\small hours-per-week} & {\small 46} & {\small 44.67} & {\small 40} & {\small 42.434} \\
    {\small native-country} & {\small United-States} & {\small United-States} & {\small United-States} & {\small United-States} \\
    {\small income} & {\small 0} & {\small 0} & {\small 1} & {\small 0} \\
    \hline
    \end{tabular}
}

\label{tab:MinDamMidDamMostDamPro}
\end{table}

%% file: 11-Appendices-german.tex
\section{Evaluation of German credit}
\label{app:germanresults}

In this appendix, we will discuss the results obtained on German credit dataset. 
First, we present the dataset distribution (Table \ref{table:datastatsgerman}).

\begin{table}[h!]
    \centering
    \caption{Distribution of the different groups with respect to the sensitive attribute and the decision one on German credit.}
    \resizebox{1\linewidth}{!}{
\begin{tabular}{|c|c|c|}
\cline{1-3}
Dataset
\multirow{2}{*}{}                         & \multicolumn{2}{c|}{\textbf{German Credit}} \\ \cline{1-3} 
                             Group             & Sensitive ($S_x=S_0$, Young) & Default ($S_x=S_1$, Old) \\ \hline
\multicolumn{1}{|c|}{$Pr(S = S_x)$}       & $19\%$                   & $81\%$ \\ \hline
\multicolumn{1}{|c|}{$Pr(Y=1 | S = S_x)$} & $57.89\%$                   & $72.83\%$ \\ \hline
\multicolumn{1}{|c|}{$Pr(Y=1)$}           & \multicolumn{2}{c|}{$70\%$} \\ \hline
\end{tabular}
}
\label{table:datastatsgerman}
\vspace{-4mm}
\end{table}

\subsection{Damage and qualitative analysis}
\label{app:germandamage}

\begin{figure*}[h!]
    \includegraphics[width=1\linewidth, height=0.2\textheight]{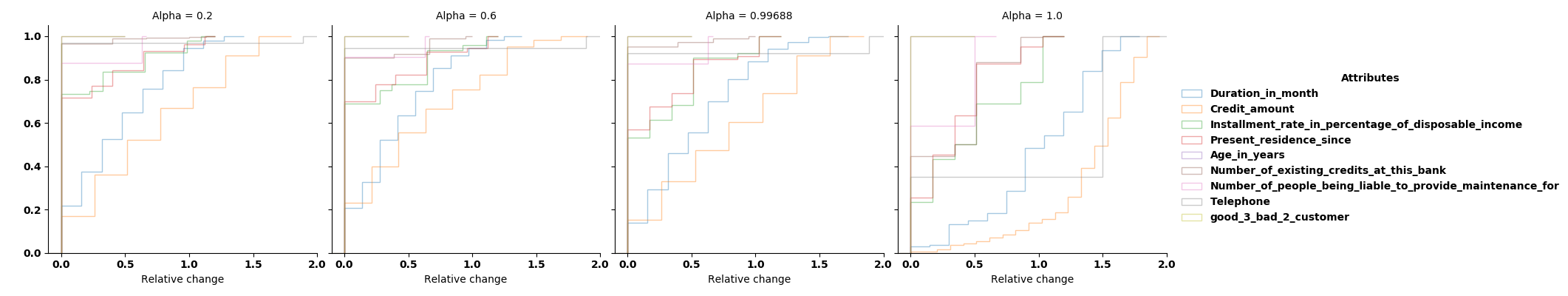}
    \caption{Relative change}
    \label{res:germanrelc}
\end{figure*}

Looking at the numerical columns damage (Figure\ref{res:germanrelc}), we can observe that most columns have a relative change lower than $0.5$ for more than $70\%$ of the dataset, regardless of the value of $\alpha$. Only columns \textit{Duration in month} and \textit{Credit amount} have a higher damage. This is due to the fact that these columns have a very large range of possible values compare to the other columns ($33$ and $921$), especially for column \textit{Credit amount} which also exhibit a nearly uniform distribution.

\begin{table}[h!]
\centering
\caption{Evaluation of \gansan{}'s protection on test. Values for reference points A, B and C}
\resizebox{\columnwidth}{!}{%
\begin{tabular}{|c|c|c|c|c|c|c|}
    \hline
    Classifier & Original & \textit{A}: $\alpha = 0.6$ & \textit{B, C}: $\alpha = 0.9968$\\\hline
    GB         & $0.3652 \pm 0.0402$ & $0.4160 \pm 0.0590$ & $0.4549 \pm 0.0411$ \\
    MLP        & $0.3723 \pm 0.0352$ & $0.3981 \pm 0.0537$ & $0.4428 \pm 0.0547$ \\
    SVM        & $0.2521 \pm 0.0434$ & $0.2868 \pm 0.0760$ & $0.3243 \pm 0.0469$ \\\hline
    
\end{tabular}
}
\label{tab:resgermantest}
\end{table}

\subsection{Evaluation scenario, other fairness metrics and utilities}
\label{app:scenariofairgerman}
\begin{figure*}[h!]
    \includegraphics[width=1\linewidth, height=0.4\textheight]{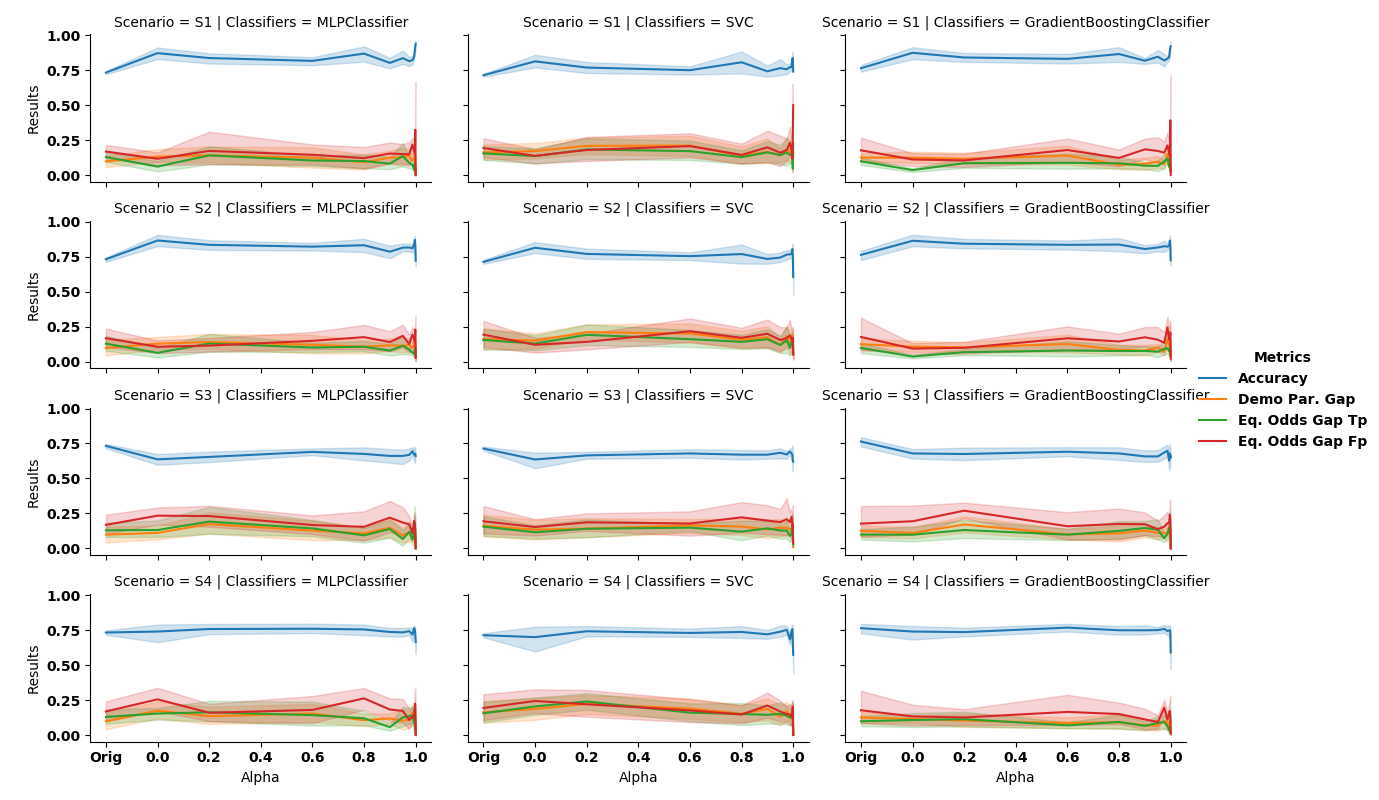}
    \caption{Accuracy (blue), demographic parity gap (orange) and equalized odds gap (true positive rate in green and false positive rate in red) computed for scenarios 1, 2, 3 and 4 (top to bottom), with the classifiers GB, MLP and SVM (left to right on german credit dataset).}
    \label{res:germanscn}
\end{figure*}

In Figure\ref{res:germanscn}, we present the results on the differents scenario investigated (\textit{S1}: complete debiasing, \textit{S2}: partial debiasing, \textit{S3}: buiding a fair classifier, \textit{S4}: local sanitization). 

First off all, we can observe that for all scenario, the accuracy is mostly stable with the increase of $\alpha$  for all classifiers. The sanitization does not significantly affect the quality of prediction, which is mostly around $75\%$, $7.143\%$ greater than the proportion of the positive outcomes in the dataset. On scenario \textit{S3} and \textit{S4} This observation comes into contrast to adult, where the accuracy decrease with the increase of the protection coefficient.

\begin{figure*}[h!]
    \includegraphics[width=1\linewidth, height=0.4\textheight]{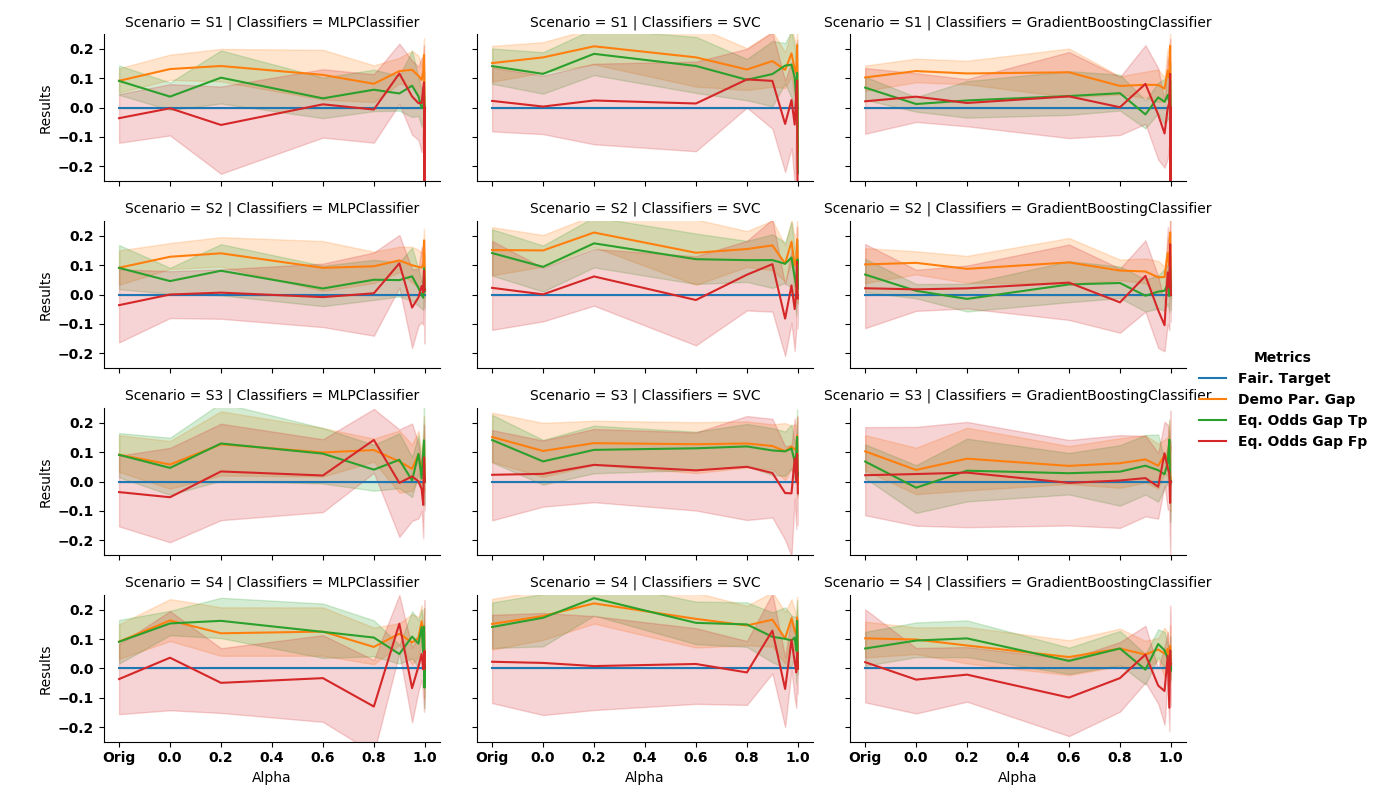}
    \caption{Fairness metrics evaluated on different scenatio on german credit dataset.}
    \label{res:germanscnzoom}
\end{figure*}

If we take a closer look at the fairness metrics as provided in Figure\ref{res:germanscnzoom}, we observe that the \myDI{} and $\eqod_1$ have a negative slope, which increase with $\alpha$. In constrast, $\eqod_0$ is rather unstable, especially when $\alpha > 0.8$. 

\paragraph{S1: complete data debiasing} In this scenario, we observe that the sanitization makes render the profiles in each decision group easily seperable, which in turn improve the accuracy as we can observe. The sanitization also reduces the risk of discrimination, just as we have seen on the Adult dataset.

\paragraph{S2: partial data debiasing}: Even though the sanitized and original decisions does not share the same distribution, the sanitization is able to transform the dataset in such way that it improves the classifications performances of all classifiers. The discrimination on the other hand is almost constant, meaning that the original decision still preserve a certain amount of discrimination that is harder to remove on not sensitive attributes alone, especially on small dataset.

\paragraph{S3: building a fair classifier}: Just as the results observed on Adult dataset, building a fair classifier by training it on sanitized data and testing/using it on unprocessed data proved to be less conservative of the accuracy.  We observe a slight drop, from $0.75$ to almost $0.65$ for the first value of $\alpha$, then it stays stable across all $\alpha$. There decision boundaries learned by the fair classifier can not be directly transfered on another type of data as they do not share the same distribution. Concerning the fairness metrics, we observe two behaviour: for $\alpha \leq 0.9$ the fairness metrics are nearly constants in constrast to adult where they all seems to increase; the discrimination is reduced when we push the system close to the maximum ($\alpha > 0.9$), but not to the extreme where the discrimination increases. The increase for extrême values is due to the fact that the sanitization has almost completely perturbed the dataset, losing all of its structure. Thus, as this set is highly imbalanced both on the sensitive attribute distribution as well as the decision one, all classifiers to predict the majority labels which are in favor of the default group (c.f. Table \ref{table:datastatsgerman}).

\paragraph{S4: local sanitization} On this dataset, this scenario provides the most significant results. As a matter of fact, the accuracy of the classifier is almost not affected by the sanitization, while the discrimination is reduced. \textit{MLP} provides the most unstable $\eqod_0$, and for all classifiers, we observe a reduction of $\myDI{}$ and $\eqod_0$, which become significantly important with higher values of $\alpha$ ($\alpha > 0.8$). This result is differents from Adult, where we observed a negative slope. An deeper analysis of such behaviour is left out as another research objective. 

Table \ref{tab:germanfairquant} provides the quantitative results for these 4 scenario on values of $\alpha$ that correspond to points \textit{A} and \textit{B}.

\begin{table*}[h!]
\centering
\caption{\gansan{} quantitative results on german credit dataset}

\resizebox{\linewidth}{!}{%
\begin{tabular}{|c|c|c|c|c|c|c|}
    \hline
    \multirow{2}{*}{$\alpha$} & \multirow{2}{*}{Classifier} & \multicolumn{5}{c|}{\textit{$\eqod_1$}} \\ \cline{3-7} 
     & & \textit{Baseline} & \textit{S1} & \textit{S2} &\textit{S3} & \textit{S4} \\ \hline
    \multirow{3}{*}{$0.6$} 
    & GB & $0.0681 \pm 0.0977$ & $0.0394 \pm 0.1271$ & $0.0345 \pm 0.1209$ & $0.0283 \pm 0.1209$ & $0.0258 \pm 0.0784$ \\ \cline{2-7}
    & MLP & $0.0910 \pm 0.1323$ & $0.0316 \pm 0.1208$ & $0.0207 \pm 0.1190$ & $0.0952 \pm 0.1448$ & $0.1249 \pm 0.1666$ \\ \cline{2-7}
    & SVM & $0.1415 \pm 0.1391$ & $0.1421 \pm 0.1488$ & $0.1207 \pm 0.1558$ & $0.1133 \pm 0.1103$ & $0.1556 \pm 0.1175$ \\ \cline{2-7}
    \hline \hline
    
    \multirow{3}{*}{$0.99688$} 
    & GB & $0.0681 \pm 0.0977$ & $0.0904 \pm 0.0960$ & $0.0871 \pm 0.0877$ & $0.0131 \pm 0.1875$ & $0.0607 \pm 0.0781$ \\ \cline{2-7}
    & MLP & $0.0910 \pm 0.1323$ & $0.0598 \pm 0.0968$ & $0.0898 \pm 0.0795$ & $0.1404 \pm 0.2049$ & $0.1425 \pm 0.0898$ \\ \cline{2-7}
    & SVM & $0.1415 \pm 0.1391$ & $0.1184 \pm 0.1593$ & $0.1012 \pm 0.1465$ & $0.1520 \pm 0.1500$ & $0.1625 \pm 0.1186$ \\ \cline{2-7}
    \hline
    
\end{tabular}
}

\resizebox{\linewidth}{!}{%
\begin{tabular}{|c|c|c|c|c|c|c|}
    \hline
    \multirow{2}{*}{$\alpha$} & \multirow{2}{*}{Classifier} & \multicolumn{5}{c|}{\textit{$\eqod_0$}} \\ \cline{3-7} 
     & & \textit{Baseline} & \textit{S1} & \textit{S2} &\textit{S3} & \textit{S4} \\ \hline
    \multirow{3}{*}{$0.6$} 
    & GB & $0.0215 \pm 0.2685$ & $0.0383 \pm 0.2317$ & $0.0407 \pm 0.2186$ & $0.0046 \pm 0.2291$ & $0.0993 \pm 0.2257$ \\ \cline{2-7}
    & MLP & $0.0362 \pm 0.2094$ & $0.0111 \pm 0.1910$ & $0.0087 \pm 0.1907$ & $0.0205 \pm 0.2038$ & $0.0330 \pm 0.2551$ \\ \cline{2-7}
    & SVM & $0.0229 \pm 0.2605$ & $0.0140 \pm 0.2635$ & $0.0186 \pm 0.2690$ & $0.0380 \pm 0.2311$ & $0.0153 \pm 0.2292$ \\ \cline{2-7}
    \hline \hline
    
    \multirow{3}{*}{$0.99688$}
    & GB & $0.0215 \pm 0.2685$ & $0.1153 \pm 0.1692$ & $0.1720 \pm 0.1889$ & $0.0730 \pm 0.2927$ & $0.0616 \pm 0.2389$ \\ \cline{2-7}
    & MLP & $0.0362 \pm 0.2094$ & $0.0862 \pm 0.1895$ & $0.0813 \pm 0.1696$ & $0.0822 \pm 0.1812$ & $0.0181 \pm 0.2855$ \\ \cline{2-7}
    & SVM & $0.0229 \pm 0.2605$ & $0.0922 \pm 0.1803$ & $0.1124 \pm 0.1813$ & $0.0944 \pm 0.1653$ & $0.0491 \pm 0.1876$ \\ \cline{2-7}
    \hline
    
\end{tabular}
}
\resizebox{\linewidth}{!}{%
\begin{tabular}{|c|c|c|c|c|c|c|}
    \hline
    \multirow{2}{*}{$\alpha$} & \multirow{2}{*}{Classifier} & \multicolumn{5}{c|}{\textit{\myDI{}}} \\ \cline{3-7} 
     & & \textit{Baseline} & \textit{S1} & \textit{S2} &\textit{S3} & \textit{S4} \\ \hline
    \multirow{3}{*}{$0.6$} 
    & GB & $0.1028 \pm 0.1048$ & $0.1204 \pm 0.1321$ & $0.1097 \pm 0.1291$ & $0.0533 \pm 0.1087$ & $0.0390 \pm 0.1034$ \\ \cline{2-7}
    & MLP & $0.0914 \pm 0.1032$ & $0.1115 \pm 0.1318$ & $0.0912 \pm 0.1400$ & $0.0991 \pm 0.1381$ & $0.1258 \pm 0.1446$ \\ \cline{2-7}
    & SVM & $0.1519 \pm 0.1456$ & $0.1715 \pm 0.1810$ & $0.1426 \pm 0.1935$ & $0.1273 \pm 0.1371$ & $0.1692 \pm 0.1573$ \\ \cline{2-7}
    \hline \hline
    
    \multirow{3}{*}{$0.99688$} 
    & GB & $0.1028 \pm 0.1048$ & $0.2109 \pm 0.1887$ & $0.2119 \pm 0.0784$ & $0.0162 \pm 0.1510$ & $0.0771 \pm 0.1161$ \\ \cline{2-7}
    & MLP & $0.0914 \pm 0.1032$ & $0.1792 \pm 0.0883$ & $0.1847 \pm 0.0709$ & $0.1290 \pm 0.1390$ & $0.1449 \pm 0.1118$ \\ \cline{2-7}
    & SVM & $0.1519 \pm 0.1456$ & $0.2130 \pm 0.0962$ & $0.1887 \pm 0.1032$ & $0.1559 \pm 0.1111$ & $0.1745 \pm 0.1145$ \\ \cline{2-7}
    \hline
    
\end{tabular}
}
\resizebox{\linewidth}{!}{%
\begin{tabular}{|c|c|c|c|c|c|c|}
    \hline
    \multirow{2}{*}{$\alpha$} & \multirow{2}{*}{Classifier} & \multicolumn{5}{c|}{\textit{\myAcc}} \\ \cline{3-7} 
     & & \textit{Baseline} & \textit{S1} & \textit{S2} &\textit{S3} & \textit{S4} \\ \hline
    \multirow{3}{*}{$0.6$} 
    & GB & $0.764 \pm 0.0566$ & $0.8311 \pm 0.0558$ & $0.0477 \pm 0.0548$ & $0.6911 \pm 0.0521$ & $0.7680 \pm 0.0483$ \\ \cline{2-7}
    & MLP & $0.733 \pm 0.0267$ & $0.8167 \pm 0.0469$ & $0.8230 \pm 0.0485$ & $0.6900 \pm 0.0412$ & $0.7600 \pm 0.0589$ \\ \cline{2-7}
    & SVM & $0.714 \pm 0.0255$ & $0.7500 \pm 0.0477$ & $0.7550 \pm 0.0477$ & $0.6789 \pm 0.0511$ & $0.7300 \pm 0.0494$ \\ \cline{2-7}
    \hline \hline
    
    \multirow{3}{*}{$0.99688$} 
    & GB & $0.764 \pm 0.0566$ & $0.8912 \pm 0.0491$ & $0.8670 \pm 0.0606$ & $0.6788 \pm 0.1123$ & $0.7500 \pm 0.0658$ \\ \cline{2-7}
    & MLP & $0.733 \pm 0.0267$ & $0.8938 \pm 0.0350$ & $0.8720 \pm 0.0533$ & $0.6725 \pm 0.0886$ & $0.7520 \pm 0.0408$ \\ \cline{2-7}
    & SVM & $0.714 \pm 0.0255$ & $0.8325 \pm 0.0373$ & $0.8060 \pm 0.0743$ & $0.6700 \pm 0.1013$ & $0.7580 \pm 0.0496$ \\ \cline{2-7}
    \hline
    
\end{tabular}
}
\label{tab:germanfairquant}
\end{table*}